\newcommand{\lga}[1]{\textcolor{red}{#1}}

\newcommand{\trans}[1]{\mathrm{trans}(#1)}
\newcommand{\Trans}[1]{\mathrm{Trans}(#1)}
\newcommand\mydots{\hbox to 0.8em{.\hss.\hss.}}

\documentclass{IOS-Book-Article}

\usepackage{mathptmx}
\usepackage{soul}\setuldepth{article}
\usepackage{xspace}
\def\hb{\hbox to 11.5 cm{}}
\usepackage{amsthm}

\newcommand{\tuple}[1]{\langle #1\rangle}

\def\Box{\mathop\square}
\def\Diamond{\mathop\lozenge}

\def\PropStandpointLogic{{\mathbb{S}}}

\def\PredStandpointLogic{{\mathbb{S}_{{\mathrm{FO}}}}}

\def\FOformulas{\PredStandpointLogic}

\def\PropConsts{\hbox{$\mathcal{P}$}\xspace}
\def\StandConsts{\hbox{$\mathcal{S}$}\xspace}
\def\PropConst#1{\hbox{$P_{#1}$}\xspace}
\def\StandConst#1{\hbox{$\s{#1}$}\xspace}

\def\PropConstSet{\{\PropConst{1}\,, \ldots, \PropConst{n} \}}
\def\StandConstSet{\{\StandConst{1}\,, \ldots, \StandConst{m}, \star \}}

\def\sp{\hbox{$\spform{s'}$}\xspace}
\def\st{\hbox{$\spform{s}$}\xspace}
\def\s#1{\hbox{$\spform{s}_{#1}$}\xspace}
\def\star{\hbox{$*$}\xspace}

\def\pr{\pi}
\def\p#1{\hbox{$\pi_{#1}$}\xspace}
\def\pp{\hbox{$\pi'$}\xspace}
\def\ppp{\hbox{$\pi''$}\xspace}

\def\PrecSet{\{\ldots, \pr_{i}\,, \ldots \}}

\newcommand{\mbf}[1]{\mathbf{#1}}
\def\Precs{\Pi}
\def\True{\mbf{t}}
\def\False{\mbf{f}}
\def\TruthValSet{\{\True, \False\}}

\def\standb#1{\Box_{\spform{#1}}}
\def\standd#1{\Diamond_{\spform{#1}}}

\def\standbs{\standb{s}}
\def\standbsp{\standb{s'}}
\def\standbi#1{\standb{s_{\spform{#1}}}}

\def\standds{\standd{s}}
\def\standdsp{\standd{s'}}
\def\standdi#1{\standd{s_{\spform{#1}}}}

\def\allstandb{\standb{*}}
\def\allstandd{\standd{*}}

\def\standindef#1{\mathcal{I}_{{\mathsf{#1}}}}
\def\allstandindef{\mathcal{I}_{*}}
\def\standdef#1{\mathcal{D}_{{\mathsf{#1}}}}
\def\allstanddef{\mathcal{D}_{*}}

\def\modelclass{\mathfrak{M}_{\PropStandpointLogic}}
\def\model{\mathfrak{M}}
\def\modelG{\model_\Gamma}

\def\modelequalstuple{\hbox{$\tuple{\Precs, \standpointFunction, \delta}\,$}}

\xspace
\def\f\xspacestandtopre{\hbox{$\sigma\,$}\xspace}
\def\fpretov\xspacealue{\hbox{$\delta\,$}\xspace}

\def\accrelationstar{\hbox{$\standpointFunction(\star)$}\xspace}
\def\accrelation#1{\hbox{$\standpointFunction(\s{#1})$}\xspace}

\def\accRelation{\hbox{$\standpointFunction$}\xspace}

\def\ModSat#1||-#2{#1\models #2}
\def\NotModSat#1||-#2{#1\nvDash #2}

\newcommand{\skipit}[1]{} 
\newcommand{\addit}[1]{} 

\newcommand{\define}[1]{\emph{#1}}
\newcommand{\N}{\mathbb{N}}
\newcommand{\set}[1]{\left\{#1\right\}}

\renewcommand{\land}{\mathrel{\wedge}}
\renewcommand{\lor}{\mathrel{\vee}}

\def\Stands{\mathcal{S}}
\def\sts{s} 
\def\Preds{\mathcal{P}}

\def\Consts{\mathcal{N}}
\def\Vars{\mathcal{X}}
\def\Terms{\mathcal{T}}
\def\E{\mathcal{E}}
\def\StandExps{\Stands}
\def\ste{s} 

\def\kstruct{\mathcal{M}}

\def\struct{\mathcal{I}}
\def\varassign{v}
\def\intf{\cdot^{\struct}}
\def\Dom{\Delta}
\def\de{\epsilon}
\newcommand{\interpret}[2]{#1^{.#2}}
\newcommand{\interprets}[1]{\interpret{#1}{\struct}}

\def\foss{first-or\-der stand\-point structure\xspace}

\usepackage{listings}
\usepackage{amsmath}
\usepackage{amssymb}
\usepackage{enumitem}
\usepackage{graphicx}
\usepackage{xcolor}
\newcommand{\pf}[1]{\mbox{\sf #1}}

\usepackage{scalerel}
\def\smallv{\scaleobj{0.75}{\mathbb{V}}}
\def\smalls{\scaleobj{0.75}{\mathbb{S}}}

\def\imp{\ \rightarrow\ }            
\def\con{\ \wedge\ }                 
\def\n{\neg}                     
\def\E{\exists}                  
\def\C{\pf{C}}

\def\A{\forall}
\def\mbf#1{\mathbf{#1}}

\def\shortcite#1{\cite{#1}}

\def\citet#1{\cite{#1}}
\def\citep#1{\cite{#1}}

\long\def\skipover#1{}

\let\mathcal\relax 
\usepackage[cal=cm]{mathalfa}

\def\VRSLogic{{\mathbb{V}_1}}

\begin{document}

\pagestyle{plain}

\begin{frontmatter}              

\title{Vagueness in Predicates and Objects}

\markboth{}{April 2022\hb}

\author[A]{\fnms{Brandon} \snm{Bennett} \orcid{0000-0001-5020-6478}} and
\author[B]{\fnms{Lucia} \snm{Gomez Alvarez}\orcid{0000-0002-2525-8839}}

\runningauthor{B.\ Bennett and L.\ Gomez Alvarez}
\address[A]{University of Leeds}
\address[B]{TU Dresden}

\begin{abstract}
Classical semantics assumes that one can model reference, predication and quantification with respect to a fixed domain of precise referent objects. 
Non-logical terms and quantification are then interpreted directly in terms of elements and subsets of this domain.
We explore  ways to generalise this classical picture of precise predicates and objects to account for variability of meaning due to factors such as vagueness, context and diversity of definitions or opinions.
Both names and predicative expressions can be given either multiple semantic referents or be associated with semantic referents that incorporate some model of variability. 
We present a semantic framework, \textit{Variable Reference Semantics},  
that can accommodate several modes of variability in relation to 
both predicates and objects.


\end{abstract}

\begin{keyword}
Vagueness\sep Semantics \sep Vague Objects \sep Standpoint Logic
\end{keyword}
\end{frontmatter}
\markboth{April 2022\hb}{April 2022\hb}

\section{Introduction}
The  notion  of  reference  and  the  operations  of  predication  and
quantification are  fundamental to  classical first-order  logic.  Its
standard semantics assumes a fixed  domain of 
 objects,  with name constants referring  to unique elements
of the  domain, predicates associated with subsets of  the domain
and quantifiers ranging over all the elements of the domain.  Thus, if
$\mathcal{D}$ is the domain of objects, then a
constant name, say  $\pf{c}$, will denote an object  $\delta(\pf{c})$, such that
$\delta(\pf{c})\in\mathcal{D}$, and a predicate $P$ will be taken to denote
a  set of  objects  $\delta(P)$, with  $\delta(P)\subseteq\mathcal{D}$.
In this way, one  can straightforwardly  interpret the  predicating expression
$P(\pf{c})$   as   a   proposition   that   is  true   if   and   only   if
$\delta(\pf{c})\in\delta(P)$.
However, in natural languages there is considerable
variability  in the  interpretation of symbols: a  name may  not
always refer  to a  unique, precisely  demarcated entity;  a predicate
need not always  correspond to a specific set  of entities. 

Building on \cite{Bennett2022jowo},
this paper explores how the classical semantics can be modified
to account for semantic variability. We shall
first  consider  some  general  ideas  regarding  the nature of vagueness
and its sources. The distinction between linguistic (\emph{de  dicto\/}) and
ontological (\emph{de re\/}) vagueness will be considered, and we will
suggest that the sources of vagueness within the interaction
between language and the world are more diverse than is
often assumed.
We  proceed to consider possible  models  of  denotation,
predication and  quantification in  the presence of  indeterminacy,
first informally, with the aid of some diagrams and then in terms of a
formal framework,  based on \emph{standpoint semantics\/} 
\cite{Bennett11lomorevi,gomez2018dealing-short,alvarez2022standpoint-short},  within which
variable  references can  be modelled.   


\section{Theories of Vagueness}

The literature on vagueness has generally assumed that
the phenomena of vagueness could arise from three potential sources
\citet{Barnes10a-short}: (a) indeterminacy of representation (linguistic a.k.a\ \emph{de dicto\/}
  vagueness), (b) indeterminacy of things in the world (known as \emph{ontic\/} 
of \emph{de
  re\/} vagueness) and (c) limitations of knowledge (\emph{epistemic\/} vagueness).
The   epistemic  view  has   some  strong   advocates
(e.g. \cite{Williamson92vague+ig-short})
and it is apparent that  semantics of multiple  possible
interpretations take a similar form to logics of knowledge and belief \cite{Lawry2009aij}. 
Indeed,      the
\emph{standpoint  semantics},  which will  be  used  in our  following
analysis can be regarded as being of this form. But the question
of whether vagueness is an epistemic phenomena, or merely shares a similar
logic, does not seem to have any direct bearing on our semantic
analysis, so  in the  current paper we  shall not further consider the epistemic view.


\subsection{Vagueness as exclusively \emph{De Dicto\/}}


A  widely held  view is  that  all vagueness  is essentially  \emph{de
dicto}. That is, it is entirely a property of language, not of the world: 
it consists in the variability or lack of precision in the way that 
words and phrases correspond to the precise entities and properties of the real world that they aim do describe.
According to this view, forms of natural language that seem to attribute vagueness directly to things in the world are misleading and can be  paraphrased in \emph{de  dicto\/} terms
\citep{Lewis86pluralityOfWorlds,Varzi01a}.  A  fairly typical version
of such  an attitude is that of  Varzi who (focusing on  the domain of
geography,  within  which  vagueness  is  pervasive)  takes  a  strong
position against  ontic vagueness.   Varzi's view of  the relationship
between vague terms and their referents is summarised in the following
quotation:
\begin{quote}
``[To]  say that  the referent  of a  geographic term  is  not sharply
  demarcated is to say that the term vaguely designates an object, not
  that it designates a vague object.'' \citep{Varzi01a} 
\end{quote}


The view that vagueness is a linguistic phenomenon
leaves open many possibilities regarding which aspects of language are involved, and how one might model vagueness within a theoretical semantic framework. These will be explored in detail below.



\subsection{The possibility of \emph{De Re\/} or \emph{Ontic\/} Vagueness}

Natural language sentences commonly describe objects as `vague': 
`The smoke  formed a vague  patch on the horizon'; 
`The vaccine injection left a vague mark on my arm'; 
`He saw the vague  outline of a  building through the  fog'.  
Here  `vague' means  something like  `indefinite in  shape,  form or
extension'. If we take
such sentences at face  value, they  seem to
indicate that vagueness can be  associated with an object in virtue of its spatio-temporal extension, its material constituency or even abstract characteristics. 
One may then argue that such properties are intrinsic to objects, and that, if
an object has a vague intrinsic property, this indicates vagueness `of
the  thing', often called \emph{de  re\/} vagueness. And, in so far as
{\em things\/} are identified with entities existing in the world, such vagueness is often described as  \emph{ontic\/} (or  \emph{ontological\/}) meaning that it pertains to what exists in reality. 

Note that the view that vagueness may be
\emph{de re\/} or \emph{ontic\/} does not require one  to deny that  vagueness is often
\emph{de  dicto}.  Even if the world consisted entirely of precisely demarcated objects with exact physical properties, there could still be vague terminology (e.g.\ adjectives such as `large') and hence vague ways to describe the world.  Proponents of \textit{de re\/} vagueness typically only claim that \textit{some\/} kinds of vagueness are ontic \citet{Tye90vague-objects,Barnes10a}.

The  idea  that vagueness  of  objects  is  primarily associated  with
vagueness of \emph{spatial extension\/} has been endorsed and examined
by Tye \citet{Tye90vague-objects}, who  gives the following criterion for
identifying vague objects: ``A concrete object $o$ is  vague if and
only if: $o$ has borderline spatio-temporal parts;
and there is no determinate fact of the matter about whether there
are objects that are neither parts, borderline parts, nor non-parts
of $o$.'' The second, rather complex condition concerns the intuition
that we cannot definitely identify borderline parts of a vague object. 
The current paper will not consider this second-order aspect of
vagueness. However, we will be presenting a semantics in which there can be objects whose spatial extension, and hence material constituents, are indeterminate. 

The case against \textit{ontic\/} vagueness is typically based on two contentions: (a) the idea of an object in the world being in itself indeterminate is mysterious and implausible; (b) statements that appear to imply \textit{ontic\/} vagueness are readily paraphrased into forms where the vagueness is evidently \emph{de  dicto}.
While it is true that, at the level of atomic 
particles, we may accept that the position of an electron might be indeterminate (as is strongly indicated by quantum mechanics), this kind of indeterminacy seems very far from what we would need to account for vagueness in the demarcation of macroscopic physical objects. Consider a pile of twigs upon a twig-strewn forest floor. The indeterminacy of whether a particular twig is part of the pile does not arise from any indeterminacy in the physical locations of twigs. Rather, it seems that, for exactly the same physical situation, there can be different judgements about which twigs should be counted as part of the pile. Nevertheless, it may still be reasonable (in accord with Tye) to take the view that the extension of the pile of twigs is indeterminate. 

So perhaps it is reasonable to consider that a `thing' such as a pile of twigs could have an indeterminate extension, even though there is no \emph{ontic\/} indeterminacy in the reality within which the twigs are manifest. Hence, if a pile of twigs is considered a `thing', the indeterminacy of its extension would be \emph{de re}, though not {\em ontic}.  Here, we see that there is an ambiguity in the meaning of `thing' (and hence in the term \emph{de re}). One can interpret it in the sense of an entity existing in a concrete reality or in the sense of an object of discourse, whose physical manifestation need not be completely determinate.

\subsection{Established Accounts of Vagueness: Fuzzy Logic and Supervauationism}

Among computer scientists  {\em fuzzy logic\/} 
\cite{Zadeh65a} 
is the dominant approach to modelling
vagueness. This theory modifies the classical
denotations of expressions.
Vague concepts denote {\em fuzzy sets\/} such that the degree of membership of an object in a fuzzy set is graded (typically represented by a real number in the range [0..1]). This degree of membership also determines fuzzy truth values in that the truth of $\C(c)$
is simply the degree of membership of the object denoted by $c$ in the fuzzy set denoted by $\C$ and the truth value of complex formulae is then determined by interpreting
logical operators in terms of fuzzy truth functions. Although the original presentation of
fuzzy logic does not treat objects as vague, 
many researchers have also modelled vague objects
as fuzzy sets of points \cite{schockaert2009spatial}.

Many philosophers favour
some  variety  of  \emph{supervaluationist\/}  account, which models linguistic
vagueness in terms of variability in the relation between vocabulary terms
and their semantic reference. This is modelled in terms of a set of
possible       precise       interpretations       (often       called
\emph{precisifications}).   An early  proposal that  vagueness  can be
analysed  in terms  of multiple  precise senses  was made  by Mehlberg
\shortcite{Mehlberg58a-short},   and   a  formal   semantics   based  on   a
multiplicity  of classical  interpretations was  used by  van Fraassen
\shortcite{Fraassen69a}  to  explain  `the logic  of  presupposition'.
This kind of formal model  was subsequently applied to the analysis of
vagueness by Fine \shortcite{Fine75a} and a similar approach was
proposed by \citet{Kamp75a}.


  
An attraction of supervaluationism is its account
\emph{penumbral connections\/}  \citep{Fine75a}, which many  believe to
be an essential characteristic of vagueness. This is
the  phenomenon  whereby  logical  laws  (such  as  the  principle  of
non-contradiction)   and   semantic   constraints  (such   as   mutual
exclusiveness  of properties such as `...  is short'  and  `... is
tall') are maintained even  for statements involving vague concepts.
The solution,  in a nutshell, being  that, even though  words may have
multiple different interpretations, each admissible precisification of
a language makes  precise all vocabulary in a  way that ensures mutual
coherence of  the interpretation of distinct  but semantically related
terms.

\section{Semantic Aspects of Predicates and Objects in Relation to Vagueness}

The  semantic  content  of  a  predicate can  be  analysed  into  many
aspects. We pick  out the following as being of particular relevance to the study of vagueness:

\begin{enumerate}[itemsep=0.5ex,labelwidth=0em,labelsep=0.5em,itemindent=0em,leftmargin=1em]
\item[]{\bf Classification:} Given a set of objects, a predicate classifies them
into members or non-members of the set. 
\item[]{\bf Individuation:} Given a relevant information about a state of the world (its spatial and material properties) the meaning of a {\em sortal\/} predicate provides
the criteria by which individual instances of the predicate are recognised. (In terms
of axioms these criteria can often be divided into {\em existential conditions\/} and 
{\em identity criteria}.)
\item[]{\bf Demarcation/Constituency:} The meaning of {\em sortal} predicates also provides
criteria for spatially demarcating and determining the material constituents
of its instances.
\end{enumerate}

In classical logic `classification' is the most prominent, 
perhaps even the only, aspect of predication.
This is because in the usual semantics for first-order logic, 
the `universe' or `domain of quantification' is typically presented as if
ontologically prior to predication, in that we  `interpret' predicates
in relation to a given domain.
However, in order to name an object or count instances of a category of object we need
to be able to individuate the object or objects to which we refer, and this implies
that individuation is necessary for (and one could say ontologically prior to) classification.
Furthermore, as illustrated by the `desert' examples below, individuating objects as instances
of a certain kind does not necessarily require that they be identified with precisely demarcated
portions of the real world. Hence, the objects of discourse referred to by names or by quantification may
not be completely definite in all their attributes.  Accounting for the independence
between criteria for individuation, classification and demarcation in the presence of vagueness is a
motivation for the semantics that we shall present below.




\subsection{Classifying, Individuating and Demarcating Deserts}\label{sec:desert-example}

\def\desertpic#1{\includegraphics[width=1.3in,angle=270,trim={40 0 40 0},clip]{#1}}

\begin{figure}
\setlength{\tabcolsep}{1mm}
\def\figwidth{1.5in}
\def\figtrim{2mm 0 0 0}
\begin{tabular}{cccc}
\desertpic{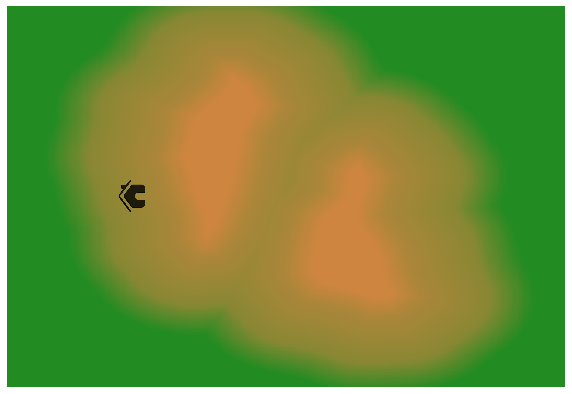} &
\desertpic{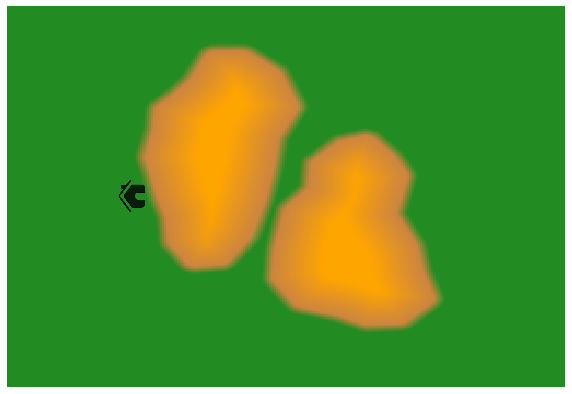} &
\desertpic{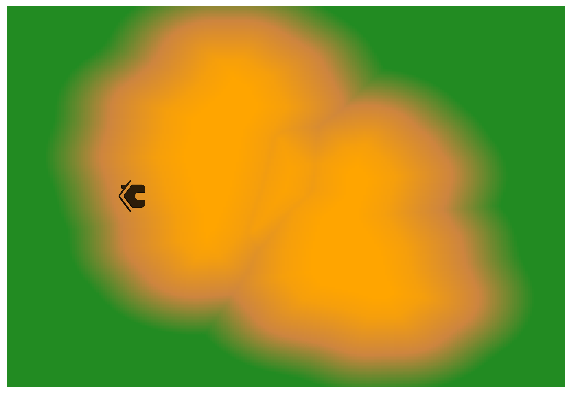} &
\desertpic{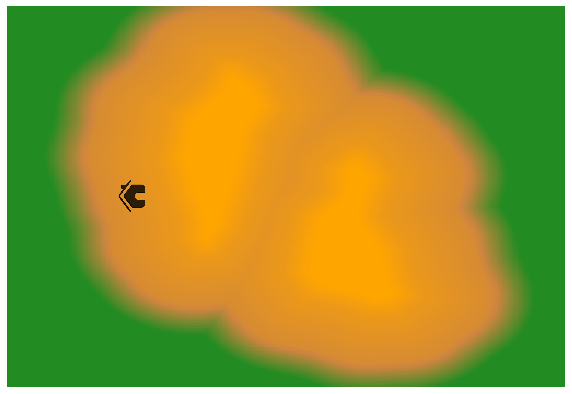} 
\\
\desertpic{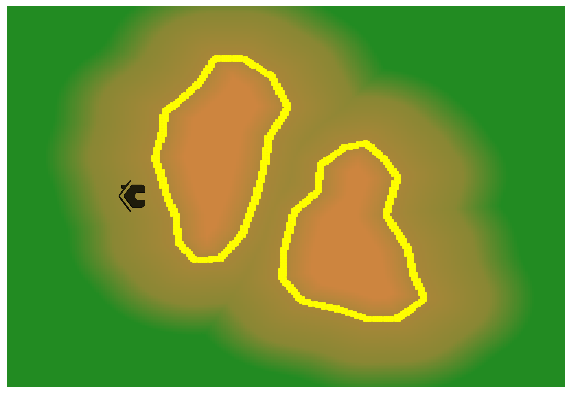} &
\desertpic{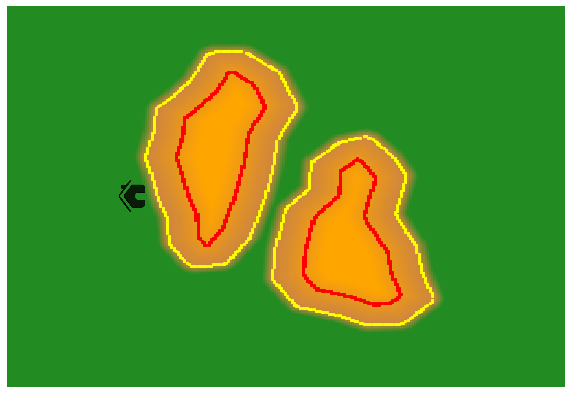} &
\desertpic{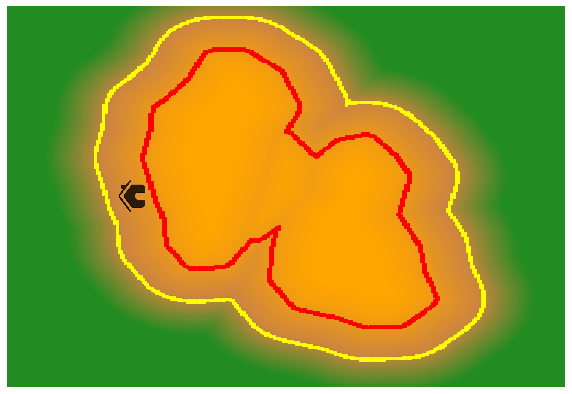} &
\desertpic{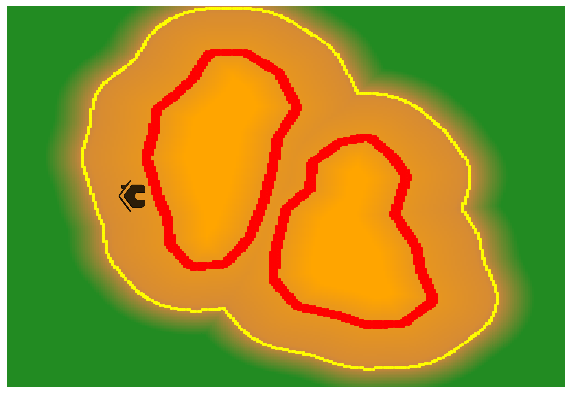} \\
(a) & (b) & (c) & (d)
\end{tabular}
\caption{Illustration of individuation, demarcation and classification
in relation to deserts.}
\label{FIG:deserts}
\end{figure}

To illustrate how the different aspects of predication operate,
we consider the concept `desert'. Several factors may be relevant to
deciding what is a desert, in particular: lack of precipitation (aridity), temperature and to a lesser extent vegetation, land form and land use. For simplicity, we consider a situation where a measure of aridity is taken to
be the only relevant factor (we may assume that other factors are either
uniform or co-vary with aridity within the region under consideration).
Consider the situations
illustrated in Fig.~\ref{FIG:deserts} depicting several states of a region with `my house'
indicated by a black icon:
\begin{itemize}
\item[(a)] The region was cool and fertile although two of the most arid areas were arguably deserts.
\emph{My house was near to one arid area that was arguably a desert.}
\item[(b)] The climate warmed and the arid areas became true deserts, but with a fertile valley between them. \emph{My house was then near to but unequivocally not part of a desert.}
\item[(c)] Eventually the desert areas expanded and merged into one big desert. \emph{Then, my house was at its boundaries, arguably though not unequivocally in the desert.}
\item[(d)] Recently a little moisture has returned along the valley. \emph{Some argue that the whole arid area is still a desert. Others refer to two deserts in that area, separated by the valley. Either way, my house is still arguably in a desert.}
\end{itemize}

The narrative sequence is not relevant to present concerns, except that it shows that situations of the same general type may exhibit different aspects of vagueness depending on specific details. 
What we should particularly note is that: in (a) the classification of the arid area as desert is equivocal;
in (b) and (c) we definitely have two deserts but each has an indeterminate boundary;
in (c) we definitely have exactly one desert, but it is debatable whether my house is within the desert area; and, in (d) the number of deserts present is indeterminate. 
The semantics that we shall develop is motivated by considering situations
such as those depicted Fig.~\ref{FIG:deserts}, where there is a separation between the aspects
of classification, individuation and demarcation with respect to vagueness. 

\subsection{Possibilities for Semantic Modelling of Vague Predicates and Objects}

Vagueness of both predicates and names can be modelled
either in terms of the mapping from vocabulary to some form of `semantic denotation'
or in the correspondence between semantic denotations and the real world. Or, indeed,
each of these relationships could model a different aspect of vagueness. 
Fig.~\ref{FIG:fuzzy+super} illustrates some possibilities.
Fig.~\ref{FIG:fuzzy+super}(a)  depicts  
a hilly region  with rocky crags, where the  terrain is
irregular  and there  is no clear  way of  dividing it  into
separate `crag' objects. The name `Arg  Crag' has been given to one of
the rocky outcrops. However, there may be different opinions regarding
exactly which outcrop is Arg  Crag. Indeed, some people  may use
the name  to refer  to the  whole of this  rocky area,  whereas others
consider that it refers to a more specific rock structure.

\def\figheight{1.2in}

\begin{figure}[htb]
\hspace{-3ex}
\centerline{
\hbox{\raise10ex\hbox{(a)}}
\includegraphics[width=1.7in]{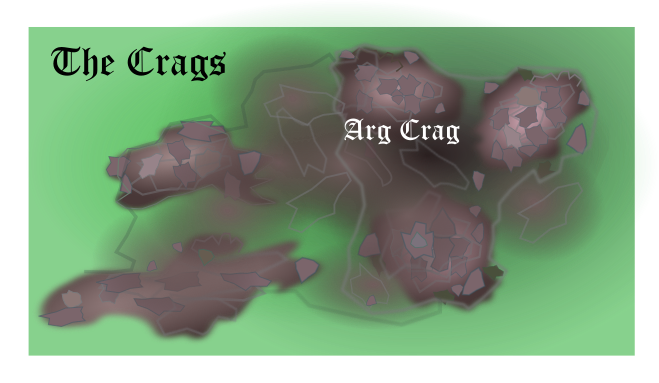}
\hbox{\raise10ex\hbox{(c)}}\hspace{-3ex}
\includegraphics[width=1.5in]{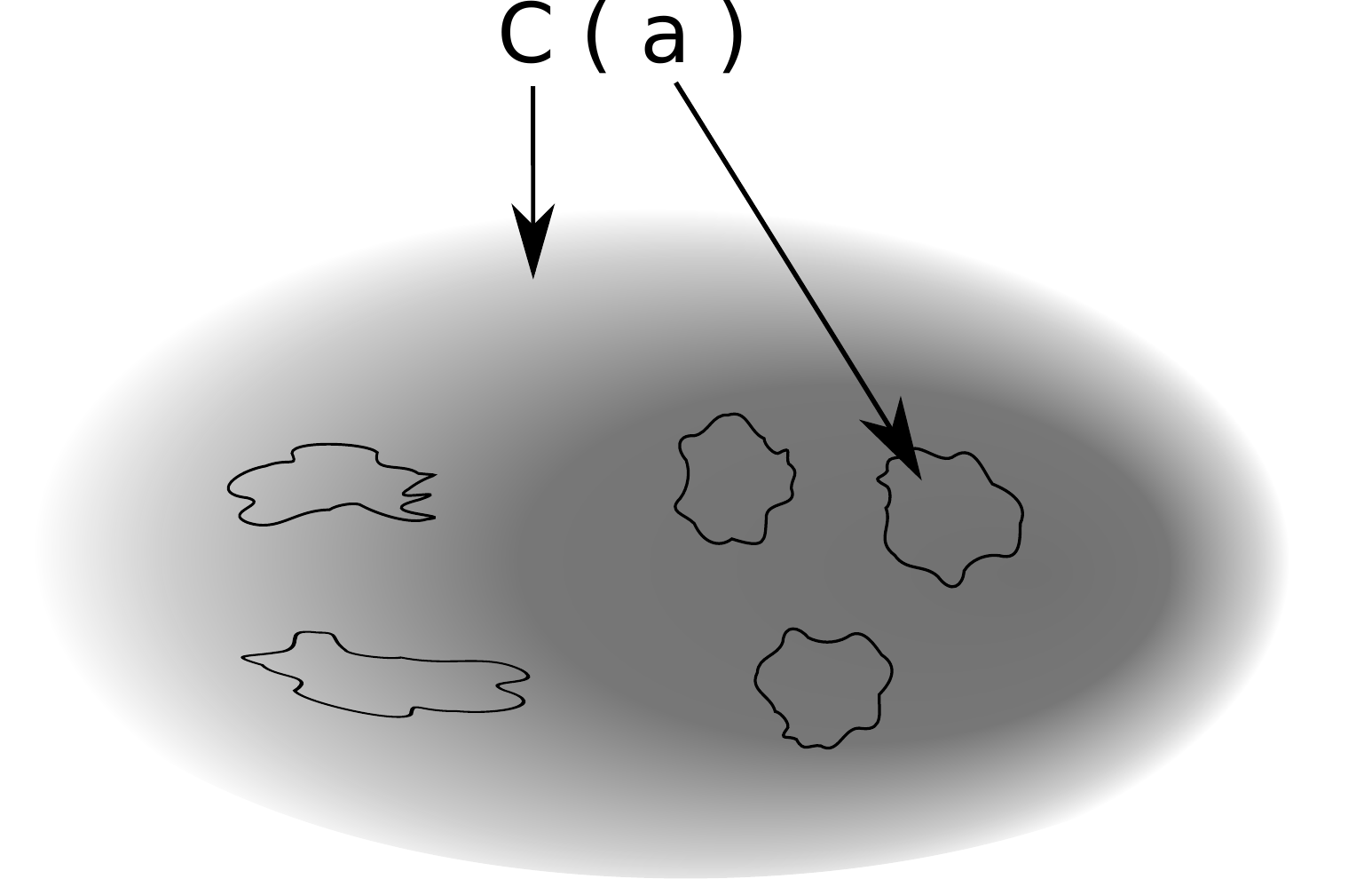}
\hbox{\raise10ex\hbox{(e)}}\hspace{-3ex}
\includegraphics[width=1.5in]{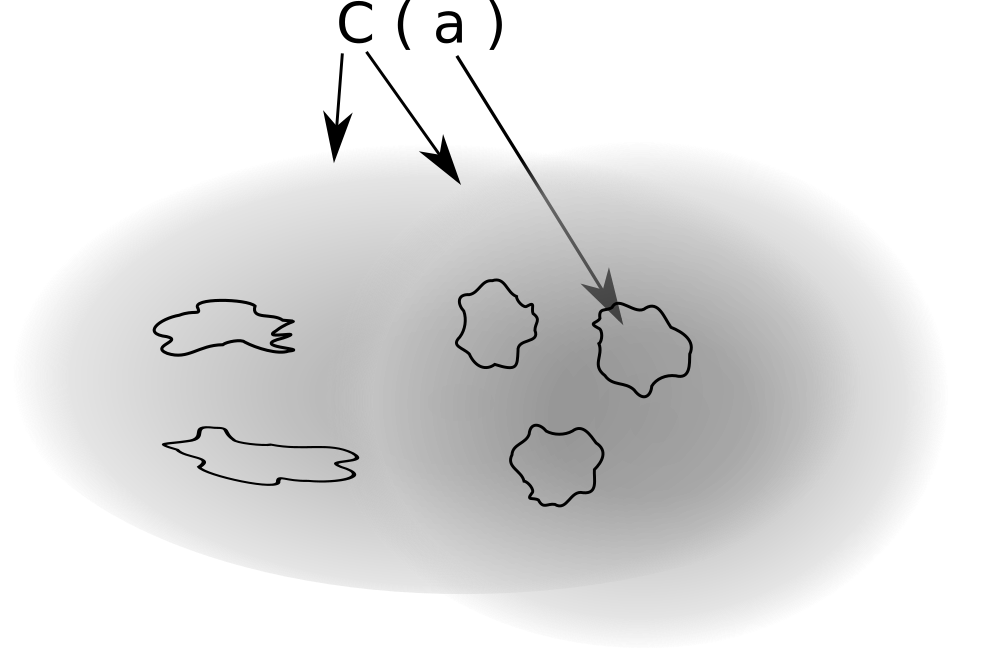}
}
\centerline{
\hbox{\raise15ex\hbox{(b)}}\hspace{-2em}
\raise3ex\hbox{\includegraphics[height=0.9in]{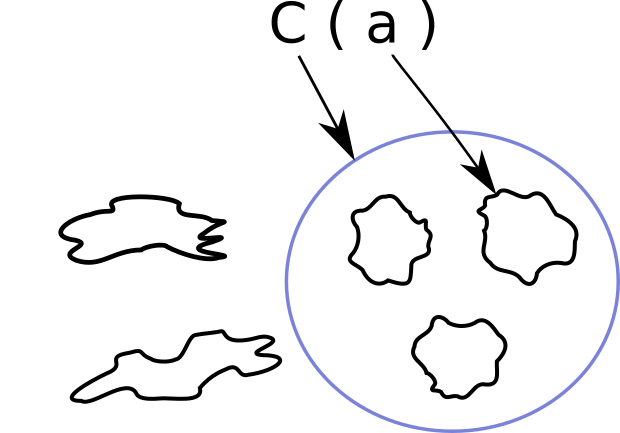}}
~~~~
\hbox{\raise15ex\hbox{(d)}}\hspace{-2em}
\includegraphics[height=\figheight]{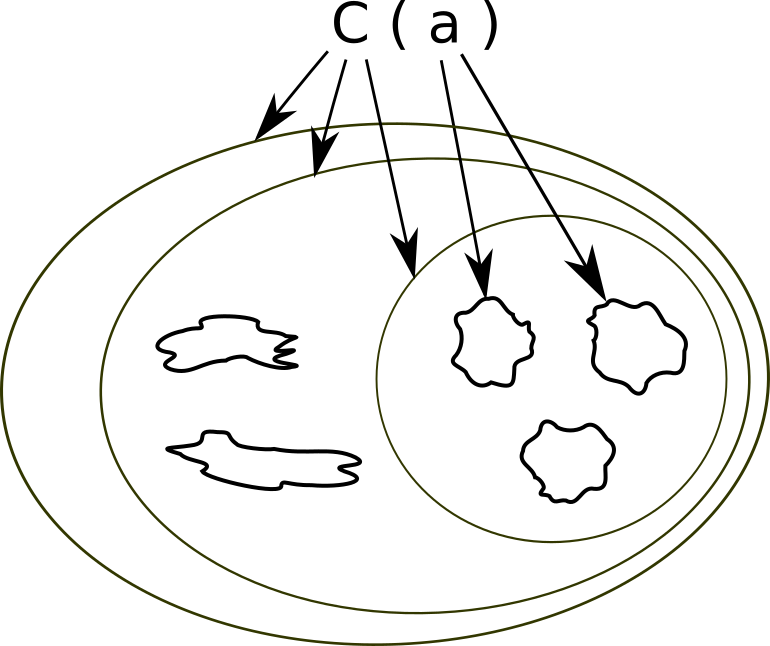}
~~~~
\hbox{\raise17ex\hbox{(f)}}\hspace{-2em}
\includegraphics[height=1.4in]{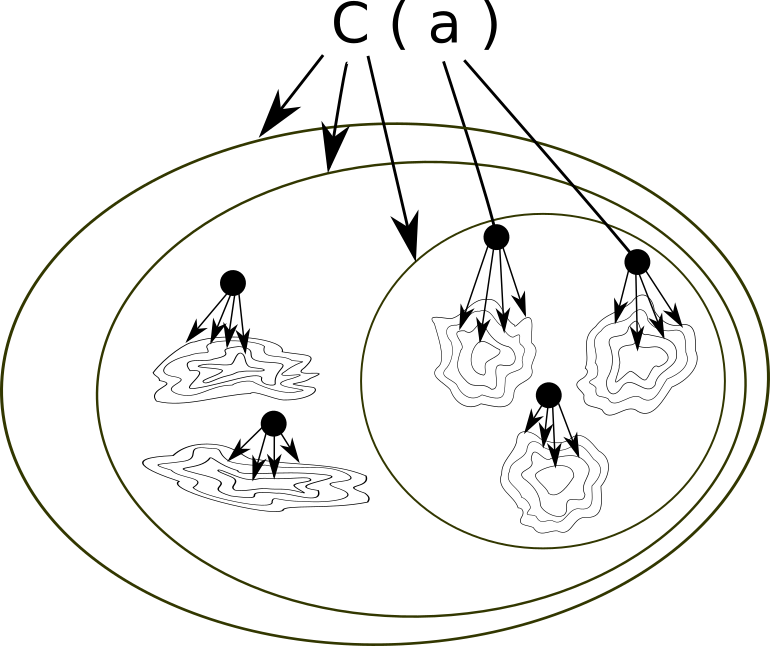}
}
\caption{
(a) Arg Crag and surrounding area. (b) Classical semantics.
(c) Fuzzy model of a vague predicate. (d) Multiple denotation
model, such as supervaluation semantics. (e) Concepts can
have multiple extensions, each of which is a fuzzy set.
(f) multiple reference for predicates and also multiple extensions for
objects.}
\label{FIG:fuzzy+super}
\end{figure}

In classical  semantics each  conceptual term  denotes a
fixed set  of entities and  each name  refers to a  single precise
entity.  Thus, even  before specifying  denotations, we  will need  to
divide the craggy  region into specific individual objects,  to make a
set of possible referents.
Figs.~\ref{FIG:fuzzy+super}(c)--(f) depict various ways in which vaguenes
of predicates and objects can be modelled. Sub-figures (c) and (d) show 
fuzzy-logic and supervaluation semantics models. 
In (c) {\sf C} denotes
a fuzzy set, whereas in (d) both the predicate and object have a variable reference
(with each {\em precisification\/} determining a unique reference).
Sub-figure (e) depicts a possible combination of the fuzzy and supervaluation approaches, where predicate symbols can have multiple references, each of which is a fuzzy set.

A limitation of models (c)--(e) is that fuzziness/variability is only associated 
with the predicate and/or the reference relation but not  the object referred to.
However, both approaches can be modified to incorporate vague
referents, that might correspond to objects with indeterminate physical boundaries.
In  fuzzy logic, objects with indefinite extensions can be modelled as 
fuzzy sets of points \cite{schockaert2009spatial}.
And within a multiple-reference semantics, one could also
associate a set of different extensions for different precise versions
of an object (see e.g.\ the
`egg-yolk'  representation  of \citet{Cohn96KRvague-short}).

Fig.~\ref{FIG:fuzzy+super}(f) 
depicts an extended form of multi-reference semantics (\emph{The Full Multi\/}).
Here we see not only that the predicates and names can have variable
reference, but also there can be multiple possible precise versions of
each reference object. This variability in the objects could
be called \emph{de re\/} vagueness, but we regard it as semantic vagueness, 
with the objects being vague semantic objects (visualised as the small black discs) that potentially correspond to many precise physical extensions. 

It may appear that the two stages of multiple
reference are redundant. The relation goes first from the name {\bf a} to one of the indefinite objects
and then to a precise extension; but surely we would get the same potential precise referents for {\sf a} if we simply assign all the precise referents we could reach as possible precise referents of {\sf a}, with no need for an intervening indefinite object. However, this would force us to conflate the
different types of vagueness illustrated in the desert examples above. So we could not have something that was definitely a desert or a mountain but whose extension was indefinite. Moreover, we believe that quantification operates in relation
to vague individuals not their precise possible extensions. This will be made explicit in our formal semantics and underpins our solution to the `Problem of the Many', which we shall discuss in Sec.~\ref{SEC:phil-views}. 

Figure~\ref{FIG:fuzzy+super}(f) still 
considerably simplifies the potential semantic variability that
might arise. It presupposes that the global set of vague
objects, together with their associations to precise entities, 
remains fixed, even though the subset associated with predicate
$\pf{C}$ may vary. In other words, $\pf{C}$ only varies in how it
\emph{classifies\/} objects, not in how it
\emph{individuates\/} them. 
The logic $\VRSLogic$ presented below is more general. It allows different
senses of sortal predicates to be associated with different ways of
individuating objects (for instance under some interpretations of
`Crag',  all three of the roundish
craggy objects within the innermost circle of the diagram might be
considered as parts of a single large crag.

\def\PropStandpointLogic{{\mathbb{S}_0}}

\def\VRSLogic{{\mathbb{V}_1}}

\def\PropConsts{\hbox{$\mathcal{P}$}\xspace}
\def\StandConsts{\hbox{$\mathcal{S}$}\xspace}
\def\PropConst#1{\hbox{$P_{#1}$}\xspace}
\def\StandConst#1{\hbox{$s_{#1}$}\xspace}
\def\Stand{\hbox{$s$}\xspace}
\def\Standp{\hbox{$s'$}\xspace}
\def\pr{\hbox{$\pi$}\xspace}
\def\p#1{\hbox{$\pi_{#1}$}\xspace}
\def\pp{\hbox{$\pi'$}\xspace}
\def\ppp{\hbox{$\pi''$}\xspace}

\def\sp{\hbox{$s'$}\xspace}
\def\st{\hbox{$s$}\xspace}
\def\s#1{\hbox{$s_{#1}$}\xspace}
\def\star{\hbox{$*$}\xspace}

\def\PropConstSet{\{\PropConst{1}\,, \ldots, \PropConst{n} \}}
\def\StandConstSet{\{\StandConst{1}\,, \ldots, \StandConst{m} \}}
\def\PrecSet{\{\ldots, \pr_{i}\,, \ldots \}}

\def\Precs{\hbox{$\Pi$}}
\def\True{\mbf{t}}
\def\False{\mbf{f}}
\def\TruthValSet{\{\True, \False\}}

\def\standb#1{\Box_{#1}}
\def\standbs{\standb{s}}
\def\standbsp{\standb{s'}}
\def\standbi#1{\standb{s_{#1}}}
\def\standd#1{\Diamond_{#1}}
\def\standds{\standd{s}}
\def\standdsp{\standd{s'}}
\def\standdi#1{\standd{s_{#1}}}
\def\allstandb{\Box_{*}}
\def\allstandd{\Diamond_{*}}
\def\standindef#1{\mathcal{I}_{#1}\,}
\def\allstandindef{\mathcal{I}_{*}\,}
\def\standdef#1{\mathcal{D}_{#1}\,}
\def\allstanddef{\mathcal{D}_{*}\,}
\def\standbi#1{\square_{s_{#1}}}
\def\standdi#1{\Diamond_{s_{#1}}}
\def\allstandb{\square_{{*}}}

\def\modelclass{\mathfrak{M}_{\PropStandpointLogic}}
\def\model{\mathcal{M}}
\def\modelG{\model_\Gamma}

\def\modelequalstuple{\hbox{$\tuple{\Precs, (\accRelation,\preceq), \delta}\,$}}
\def\accRelation{\hbox{$S$}\xspace}
\def\accrelation#1{\hbox{$\s{#1}$}\xspace}
\def\accrelationstar{\hbox{$\star$}\xspace}

\def\deltaP{\delta_{\mathcal{A}}}
\def\deltaK{\delta_{\mathcal{K}}}
\def\deltaQ{\delta_{\mathcal{Q}}}
\def\deltaR{\delta_{\mathcal{R}}}
\def\deltaS{\delta_{\mathcal{S}}}
\def\deltaT{\delta_{\mathcal{T}}}
\def\deltaN{\delta_{\mathcal{N}}}
\def\deltaX{\delta_{\mathcal{X}}}
\def\deltaV{\delta_{\mathcal{V}}}
\def\deltaF{\delta_{\mathcal{F}}}
\def\deltaG{\delta_{\mathcal{G}}}
\def\deltaY{\delta_{\mathcal{Y}}}

\section{A Semantic Theory of Variable Reference}

We now  consider what kind  of semantics  can account for  the general
form       of      variable       denotation      illustrated       in
Fig.~\ref{FIG:fuzzy+super}(f).  \emph{Standpoint Semantics\/} provides a
reasonably general framework within which 
semantic variability can be modelled in terms
of  the  symbols  of  a   formal  language  having  multiple  possible
denotations. We shall start by introducing first-order standpoint logic
 and then elaborate this to the formalism
\emph{Variable  Reference  Logic},  within   which  we  can  model
predication and quantification of vague objects. 

\subsection{Standpoint Logic}
Standpoint Semantics is based on a formal structure that models
semantic variability in terms of the following two closely connected
components:

\begin{itemize}
\item A \emph{precisification\/} is a precise and consistent interpretation of a vague
  language, and it coherently assigns precise denotations to all its
  vague elements. A model for a vague sentence
  contains a set of admissible precisifications of that sentence.
\item A \emph{standpoint\/} is modelled as a set of
  \emph{precisifications} which are compatible with a
  point of view or context of language understanding. 
  A standpoint can capture both
  explicit specifications of terminology and implicit
  constraints on meanings that arise in conversation 
  (e.g. ``That person is tall'' constrains the meaning of tallness).
\end{itemize}

\subsubsection{Syntax of First-Order Standpoint Logic}
First-order standpoint logic ($\FOformulas$) 
 is based on a \define{vocabulary} $\tuple{\Preds, \Consts, \Stands}$, consisting of
 \define{predicate symbols} $\Preds$ 
 (each associated with an arity \mbox{$n\in\N$}), 
 \define{constant symbols} $\Consts$ and \define{standpoint symbols} 
 $\Stands$, usually denoted with 
 $\sts,\sts'$, such that $\star\in\Stands$, where $\star$ is used to designate the \emph{universal
  standpoint}. 
  There is also a set $\Vars=\{ x, y, \mydots\}$ of \define{variables}, 
  and the set $\Terms=\Consts\cup\Vars$ of \define{terms} contains all constants and variables. 
  The set $\FOformulas$ of first-order standpoint \define{formulae} is given by
  \\[1ex]
\centerline{	
 $ \phi,\psi := \pf{P}(t_1,\dots,t_k) \mid \n\phi \mid \phi\con\psi \mid \forall x\phi \mid \standb{s}\phi\ , $
 }\\[1.2ex]
	where \mbox{$\pf{P}\in\Preds$} is a $k$-ary predicate symbol, \mbox{$t_1,\dots,t_k\in\Terms$} are terms,
	\mbox{$x\in\Vars$}, and \mbox{$\ste\in\StandExps$}.
The definable connectives and operators $\mathbf{t}$, $\mathbf{f}$, \mbox{$\phi\lor\psi$}, \mbox{$\phi\rightarrow\psi$}, \mbox{$\exists x\phi$}, and \mbox{$\standd{\ste}\phi$} are  as usual.


\subsubsection{Semantics of First-Order Standpoint Logic}\label{sec:sem-standpoint-logic}

Given a vocabulary $\tuple{\Preds,\Consts,\Stands}$, a \define{\foss} $\kstruct$ is a tuple $\tuple{\Dom, \Precs, \sigma, \delta}$, where 
		$\Dom$ is the (non-empty) 
		\define{domain} of $\kstruct$,
		$\Precs$ is a non-empty set of \define{precisifications},
		$\sigma$ is a function mapping each standpoint symbol from $\Stands$ to a set of precisifications (i.e., a subset of $\Precs$), and
		$\delta$ is a function mapping each precisification from $\Precs$ to an ordinary first-order structure $\struct$ over the domain $\Delta$, whose interpretation function $\intf$ maps\/:

\vspace{-2ex}
  
		      \begin{itemize}
			      \item each predicate symbol $\pf{P}{\,\in\,}\Preds$ of arity $k$ to an $k$-ary relation \mbox{$\interprets{\pf{P}} {\,\subseteq\,} \Dom^k$},
			      \item each constant symbol $a{\,\in\,}\Consts$ to a domain element \mbox{$\interprets{a}{\,\in\,}\Dom$}.
		      \end{itemize}
        
The semantics of \cite{gomez-alvarez22howtoagree-short} 
required that $\interpret{a}{\delta(\pr_1)}=\interpret{a}{\delta(\pr_2)}$ for any  $\pr_1,\pr_2\in\Precs$ and 
$a\in\Consts$. Here we allow \define{non-rigid constants} that  may denote different objects in different precisifications.
But (in contrast to the more expressive $\VRSLogic$ that will be developed below) all precisifications of a first-order standpoint structure will implicitly share the same interpretation domain $\Dom$; that is, the \emph{constant domain assumption} is adopted.
The most distinctive elements of  the model are the $\accrelation{i}$,
which model  the notion  of standpoint as a set  of precisifications
that are \emph{admissible\/} for that standpoint. 
A proposition is {\em unequivocally true\/} according
to standpoint $\s{i}$ iff it is true at every precisification $\pr\in \s{i}$.

	Let $\kstruct=\tuple{\Dom,\Precs,\sigma,\delta}$ be a \foss for the vocabulary $\tuple{\Preds,\Consts,\Stands}$ and $\Vars$ be a set of variables.
	%
	%
	A \define{variable assignment} is a function $\varassign:\Vars\to\Dom$ mapping variables to domain elements.
	Given a variable assignment $v$, we denote by $\varassign_{\set{x\mapsto\de}}$ the function mapping $x$ to $\de\in\Dom$ and any other variable $y$ to $\varassign(y)$.

\noindent
	An interpretation function $\intf$ and a variable assignment specify how to interpret terms as domain elements\/:\footnote{ Here, and in the following, $\interpret{\sigma}{\gamma}$ 
 is a concise notation for $\gamma(\sigma)$.
 It gives the semantic denotation of symbol $\sigma$ according to a function $\gamma$.}
 
 \vspace{-5ex}
	\begin{gather*}
		\interpret{t}{\struct,\varassign} =
		\begin{cases}
			\varassign(x)  & \text{if } t=x\in\Vars,   \\
			\interprets{a} & \text{if } t=a\in\Consts.
		\end{cases}
	\end{gather*}

\vspace{-0.5ex}

\noindent Then,	let $\pr\in\Precs$ and $\varassign:\Vars\to\Dom$ be a variable assignment.
	The satisfaction relation $\models$ is:%
	\vspace{-1ex}
	\addtolength{\jot}{-0.4em}
	\begin{align*}
		 & \kstruct,\!\pr,\!\varassign\ \models\ \pf{P}(t_1,\ldots,t_k)\!\! & \text{iff~~}\ \  & (\interpret{t_1}{\delta(\pr), \varassign},\ldots,\interpret{t_k}{\delta(\pr),\varassign}) \in \interpret{\pf{P}}{\delta(\pr)} \\
		 & \kstruct,\!\pr,\!\varassign\ \models\ \neg\phi                     & \text{iff~~}\ \  & \kstruct,\!\pr,\!\varassign\not\models\phi                                                                                      \\
		 & \kstruct,\!\pr,\!\varassign\ \models\  \phi\land\psi                & \text{iff~~}\ \  & \kstruct,\!\pr,\!\varassign\ \models\ \phi \text{ and } \kstruct,\pr,\varassign\models\psi                                          \\
		 & \kstruct,\!\pr,\!\varassign\ \models\  \forall x\phi                & \text{iff~~}\ \  & \kstruct,\!\pr,\!\varassign_{\set{x\mapsto\de}}\ \models\ \phi \text{ for all } \de{\,\in\,}\Dom                                    \\
		 & \kstruct,\!\pr,\!\varassign\ \models\ \standb{\ste}\phi            & \text{iff~~}\ \  & \kstruct,\!\pr'\!,\varassign\ \models\ \phi \text{ for all } \pr'\in\sigma(\ste)                                                  \\
		 & ~~~\kstruct,\!\pr\ \models\ \phi                                      & \text{iff~~}\ \  & \kstruct,\!\pr,\!\varassign\ \models\ \phi \text{ for all } \varassign:\Vars\to\Dom                                                 \\
		 & ~~~~~~ \,\kstruct\ \models\ \phi                & \text{iff~~}\ \  & \kstruct,\!\pr\ \models\ \phi \text{ for all } \pr\in\Precs
	\end{align*}

\noindent When \mbox{$\kstruct\models\phi$} we say that $\kstruct$ is a \define{model} for $\phi$. 




\vspace{1ex}

The logic $\FOformulas$ enables one to formalise the content of some statements discussed in Section \ref{sec:desert-example}, namely (1), ``\emph{My house was near to one arid area that was arguably a desert}'' and (2), ``\emph{My house was then near to but unequivocally not part of a desert}''. We assume that $\pf{Desert}$ and $\pf{AridArea}$ are vague predicates, but (my) $\pf{house}$ has definite coordinates and $\pf{PartOf}$ is a definite predicate.

\vspace{-2em}
\begin{gather}
\exists\, a\ \Box_{\star}(\pf{AridArea}(a)\con \pf{Near}(\pf{house},\pf{a})) \con \Diamond_{\star}\pf{Desert}(\pf{a})  \tag{a}  \\
\Box_{\star}\exists\, a\ (\pf{AridArea}(a)\con \pf{Near}(\pf{house},\pf{a}) \con \neg\pf{PartOf}(\pf{house},\pf{a}) \con \pf{Desert}(\pf{a}))\tag{b}
\end{gather}

\vspace{-0.5em}
It should however be apparent that addressing examples involving vague objects with $\FOformulas$ becomes less straightforward and would involve, at least, a reasonable amount of paraphrasing. For instance, one may be inclined to formalise sentence (3) as 

\vspace{-2em}
\begin{gather}
\exists\, a\ \Box_{\star}(\pf{AridArea}(a)\!\con\! \pf{Desert}(a))\! \con\! \Diamond_{\star}\pf{PartOf}(\pf{house},\pf{a})\! \con\! \neg\Box_{\star}\pf{PartOf}(\pf{house},\pf{a})\tag{c} 
\end{gather}

\vspace{-0.5ex}

However, if the domain only contains precise individuals (e.g. all possible areas), each specific area $a$ has the same extension in all precisifications, so  $\pf{house}$ cannot be arguably but not definitely part of it. Hence, this formula is unsatsifiable in $\FOformulas$, unless \pf{PartOf} can be indefinite, which is 
contrary to its intended (purely geometrical) meaning.

\subsection{Variable Reference Logic}
\label{SEC:VRL}

We now generalise standpoint semantics  to 
define variable reference logic, $\VRSLogic$,
that can represent predication and quantification over indefinite objects.
The most significant complication of $\VRSLogic$ in relation to $\FOformulas$
is that rather than the domain being comprised only of atomic entities, we also introduce
`indefinite individuals' that are modelled as functions from precisifications
to precise entities. Another complication is that predicates are divided into:
`sortals' (corresponding to count nouns such as `desert' or `cat') 
that determine subdomains of the domain of indefinite individuals,
`indefinite predicates' that classify or relate indefinite individuals, and
`precise entity predicates' that represent properties or relations of exact spatio-temporal
and/or material extensions in reality. 

With respect to sortals, individual properties and quantification, indefinite individuals are treated as basic objects (e.g.\ a particular mountain or fluffy cat). But in the application of a `precise entity predicate'  (e.g.\ `... contains $7.3{\times}10^{26}$ atoms') the individual is evaluated with respect to a precisification to obtain a precise entity (the individual's extension) that determines the truth of the precise predicate. Although our semantics allows indefinite individuals to be any function from precisifications to precise entities, instances of sortals will typically by highly constrained by
axioms expressing conditions arising from the meaning of the sortal. For example, if a desert is a maximal area of the Earth's surface satisfying certain conditions (e.g.\ low precipitation) then, even if exact thresholds on the parameters of these conditions are not given, the set of indefinite individuals that could correspond to a desert is hugely restricted.

\subsubsection{Syntax}

The language of $\VRSLogic$ is built from a vocabulary
$\mathcal{V} =
\tuple{\mathcal{K},\mathcal{A},\mathcal{Q},\mathcal{N},
\mathcal{S}}$,
comprising:
\begin{itemize}[leftmargin=2em, itemsep=0ex,label={~$\bullet$~}]
   \item $\mathcal{K} = \{ \ldots, \mathsf{K}_i, \ldots \}$, ~~ sortal predicates\ (e.g.\ \pf{Desert}, \pf{Mountain}, \pf{Cat}),
   \item $\mathcal{A} = \{ \ldots, \mathsf{A}_i, \ldots \}$, ~~ indefinite predicates\ (e.g.\ \pf{Dry}, \pf{Fluffy}, \pf{Climbed}),
   \item $\mathcal{Q} = \{ \ldots, \mathsf{Q}_i, \ldots \}$, ~~ precise entity predicates\ 
    (e.g.\ exact spatial properties),
   \item $\mathcal{N} = \{ \ldots, \mathsf{n}_i, \ldots \}$, ~~ proper name symbols\ 
   (e.g.\ \pf{everest}, \pf{tibbles}).
   \item $\mathcal{S}\, = \{ \ldots, s_i, \ldots, \star \}$, ~~ standpoint symbols.
\end{itemize}

Let $\Preds=\mathcal{K}\cup\mathcal{A}\cup\mathcal{Q}$ be the set of predicate symbols, containing all sortals, indefinite predicates and precise entity predicates. 
Again, there is a set of variables $\mathcal{X}$, and the set $\mathcal{T}= \mathcal{N}\cup\mathcal{X}$ contains all proper name symbols and variables.
Then, the set $\VRSLogic$ of variable reference logic \define{formulae} is given by\\[1ex]
\centerline{$ 
\phi,\psi \ \ := \ \ \pf{P}(\tau) \ \mid\ \n\phi \ \mid\ \phi\!\con\!\psi \ \mid\ \forall x\, \phi \ \mid\ \standb{s}\phi \ , 
$}
\\[1.2ex]
where \mbox{$\pf{P}\in\Preds$}, \mbox{$\tau\in\Terms$},
	\mbox{$x\in\Vars$}, \mbox{$\pf{K}\in\mathcal{K}$} and \mbox{$\ste\in\StandExps$}.

\subsubsection{Semantics}

Given a vocabulary $\mathcal{V}= \tuple{ \mathcal{K}, \mathcal{A}, \mathcal{Q},
  \mathcal{N}, \mathcal{S} }$, a \emph{variable reference logic structure} is a tuple
$\tuple{ E, \Pi, \sigma, \delta}$
where:
\begin{enumerate}[label=$\bullet$,itemsep=0ex]
\item $E$\ is the set of precise entities.
\item $\Pi$\ is the set of precisifications.
\item $\sigma$\  
   is a function mapping standpoint symbols to precisifications, and
\item $\delta = \tuple{  
         \delta_{\mathcal{K}},
         \delta_{\mathcal{A}},
         \delta_{\mathcal{Q}},
         \delta_{\mathcal{N}}
         }$ is a denotation function divided into components specifying the denotations for
  each type of non-logical symbol (see below).
\end{enumerate}

\def\idom{I_{*}}



We define the set of \textit{indefinite individuals} $\idom=Maps(\Pi,E)$ as the set of all mappings from  precisifications to precise entities. 
For each indefinite individual $i\in \idom$ and $\pi\in\Pi$,  $i(\pi)$ will be an element of $E$, which constitutes the precise version of individual $i$ according to precisification $\pi$. 
At each precisification $\pi\in\Pi$, each name symbol $\mathsf{n}\in\mathcal{N}$ denotes some element of $\idom$
and each sortal $\mathsf{K}\in\mathcal{K}$ and individual attribute predicate $\mathsf{A}\in\mathcal{A}$ denotes a subset of $\idom$.
The set of indefinite individuals at a precisification $\pi$ depends on the interpretation of the sortal predicates at $\pi$. The denotation functions map:
\begin{enumerate}[itemsep=1ex,label=$\bullet$]
\item $\delta_{\mathcal{K}} : $  each precisification $\pi\in\Pi$ into a function from each sortal symbol $\pf{K}{\,\in\,}\mathcal{K}$ to a subset  $\interpret{\pf{K}}{\delta_{\mathcal{K}}(\pi)}\subseteq\idom$ --- i.e.\ a set of indefinite individuals.
\\[0.5ex]
On the basis of $\delta_{\mathcal{K}}$ we define
     $I_{\pr} =  \bigcup \{ \interpret{\mathsf{K}}{{\delta_{\mathcal{K}}(\pi)}} \ |
  \ \mathsf{K} \in \mathcal{K}\}$, the domain of (indefinite) individuals according to
  precisification $\pi$. 
 $I_{\pr}$ is
the set of all individuals of any sort, where the sortals are interpreted according to precisification $\pr$.

\item $\delta_{\mathcal{A}} : $ each precisification $\pi\in\Pi$ into a function from each indefinite predicate symbol $\pf{A}{\,\in\,}\mathcal{A}$ of arity $k$ to a $k$-ary relation $\interpret{\pf{A}}{\delta_{\mathcal{A}}} {\,\subseteq\,} I_{\pi}^k$,

\item $\delta_{\mathcal{Q}} : $ each entity predicate symbol $\pf{Q}{\,\in\,}\mathcal{Q}$ of arity $k$ to $k$-ary relations $\interpret{\pf{Q}}{\delta_{\mathcal{Q}}} {\,\subseteq\,} E^k$,

\item $\delta_{\mathcal{N}} : $ each precisification $\pi\in\Pi$ into a function from each indefinite name symbol $\pf{N}{\,\in\,}\mathcal{N}$ to an indefinite individual $\interpret{\mathsf{N}}{{\delta_{\mathcal{N}}(\pi)}}\in I_\pi$.

\end{enumerate}

Let us now deal with the variable assignments in \emph{variable reference logic}. Let $\mathcal{M} =\tuple{ E, \Pi, \sigma, \delta }$ be a model for the vocabulary $\tuple{ \mathcal{K}, \mathcal{A}, \mathcal{Q}, \mathcal{N}, \mathcal{S} }$ and $\Vars$ be a set of variables.
A \define{variable assignment} is a function $\varassign:\Vars\to\idom$ mapping variables to individuals.
Given a variable assignment $v$, we denote by $\varassign_{\set{x\mapsto i}}$ the function mapping $x$ to $i \in\idom$ and any other variable $y$ to $\varassign(y)$.

The interpretation function $\delta_{\tiny\mathcal{N}}$ and a variable assignment specify how to interpret terms by domain elements in each precisification $\pi\in\Pi$\/:
\vspace{-1ex}
\begin{gather*}
	\interpret{\tau}{\pi,\varassign} =
	\begin{cases}
		\varassign(x)  & \text{if } \tau=x\in\Vars,   \\
		\interpret{\pf{n}}{\delta_{\mathcal{N}}(\pi)} & \text{if } \tau=\mathsf{n}\in\Consts.
	\end{cases}
\end{gather*}

\noindent Then,	let $\pr\in\Precs$ and $\varassign:\Vars\to\Dom$ be a variable assignment.
The satisfaction relation $\models$ is:%

\vspace{-3ex}
	\begin{align*}
		 & \kstruct,\!\pr,\!\varassign \models \pf{K}(\tau)\!\! & \text{iff}\ \  & \interpret{\tau}{\pi,\varassign} \in \interpret{\pf{K}}{\deltaK(\pi)}\\
		 & \kstruct,\!\pr,\!\varassign  \models \pf{A}(\tau_1,\ldots,\tau_k)\!\! & \text{iff}\ \  & \langle\interpret{\tau_1}{\delta(\pr), \varassign},\ldots,\interpret{\tau_k}{\delta(\pr),\varassign}\rangle \in \interpret{\pf{A}}{\deltaP(\pi)}\\
		 & \kstruct,\!\pr,\!\varassign  \models \pf{Q}(\tau_1,\ldots,\tau_k)\!\! & \text{iff}\ \  & \langle\interpret{\tau_1}{\delta(\pr), \varassign}(\pi),\ldots,\interpret{\tau_k}{\delta(\pr),\varassign}(\pi)\rangle\in \interpret{\pf{Q}}{\deltaQ} \\
		 & \kstruct,\!\pr,\!\varassign \models \neg\phi                     & \text{iff}\ \  & \kstruct,\!\pr,\!\varassign\not\models\phi                                                                                      \\
		 & \kstruct,\!\pr,\!\varassign \models \phi\land\psi                & \text{iff}\ \  & \kstruct,\!\pr,\!\varassign\models\phi \text{ and } \kstruct,\pr,\varassign\models\psi                                          \\
		 &\kstruct,\!\pr,\!\varassign \models \A x\,\phi(x)    & \text{iff}\ \  & \kstruct,\!\pr,\!\varassign_{\set{x\mapsto i}} \models\  \phi(x)
\hbox{~~for all}\ i \in I_\pi                                  \\
		 & \kstruct,\!\pr,\!\varassign \models \standb{\ste}\phi            & \text{iff}\ \  & \kstruct,\!\pr'\!,\varassign \models \phi \text{ for all } \pr'\in\sigma(\ste)                     
	\end{align*}

To understand the interpretation of atomic
predications, one must be  aware  that the
evaluation  of a  symbol  may  require
one  or  two levels  of
de-referencing in  relation to  the precisification index  $\pi$.  First, notice
that the  $\pf{K}$ and  $\pf{A}$ predications are interpreted in the same way.   When applied to a name symbol  {\sf n},
both the  constant and the predicate  get evaluated with respect  to a
precisification, so that the names denote particular individuals and
the predicate denotes a set of such individuals.  When the argument is
a variable rather than a  name symbol, the variable directly denotes
an  individual  without   any  need  for  evaluation   relative  to  a
precisification.   In  the  case  of  (exact)  $\pf{Q}$  predications,
individuals   need   to  be   further   evaluated   relative  to   the
precisification  in order  to obtain precise  entities, which  can be
tested  for  membership  of  the   precise  set  denoted  by the property
$\pf{Q}$. So,  although $\pf{Q}$ predicates are  not themselves subject
to  variation in  relation  to  $\pi$, 
they require  an  extra level  of
disambiguation in the interpretation of their argument symbol. 

\subsubsection{Classification, Individuation and Demarcation with $\VRSLogic$}

The elaboration in the denotation functions for the different kinds
of symbol  and the semantics  given above  for the different  cases of
predication allow quantification to operate at an
intermediate level in the  interpretation of
reference.  The  \emph{individuation\/} of potential  referents occurs
prior to quantification  by establishing individuals in  relation to a
given interpretation  of sortal predicates. But  these individuals can
still  be indeterminate  in that  they  may correspond  to different  exact
entities.
The formulas (a) and (b) in Section \ref{sec:sem-standpoint-logic} are also satisfiable $\VRSLogic$ formulas, and correctly formalise the scenarios described. 
Furthermore, the formula (c), that was unsatisfiable in $\FOformulas$,
is satisfiable in $\VRSLogic$, because area is modelled as an indefinite individual 
that is unequivocally classified as a $\pf{Desert}$ and at some but not all precisifications is associated to a precise extension containing $\pf{house}$.
Finally, statement (d), ``\emph{Some argue that the whole arid area is still a desert. Others refer to two deserts in that area, separated by the valley. Either way, my house is still arguably in a desert}'', can be formalised as follows:\\[-3ex]
\begin{gather*}
    \exists\, a\ (\Box_{\star}\pf{AridArea}(a)\con \Diamond_{some}\pf{Desert}(a) \\
    \con \Diamond_{others}(\exists\, b,c\ (b \neq c) \con \pf{PartOf}(b,a) \con \pf{PartOf}(c,a) \con \pf{Desert}(b)\con \pf{Desert}(c)))\\
    \con \Diamond_{*}\exists\, x\ \pf{Desert}(a) \con  \pf{PartOf}(\pf{house},x)
\end{gather*}

\vspace{-1ex}


Thus, all statements in our example involving the classification, individuation and demarcation of vague objects can be easily represented in $\VRSLogic$.

\section{From $\VRSLogic$ back to First Order Standpoint Logic}


We next turn to the question of the expressive power of $\VRSLogic$, specifically in comparison to $\FOformulas$.
We will provide a translation from $\VRSLogic$ into $\FOformulas$, which not only settles the above question but is also interesting in its own right, since it spells out the restrictions that $\VRSLogic$ imposes on the semantics and facilitates the discussion on some ontological commitments that are arguably implicit when favouring the {\em de dicto\/} translation of a $\VRSLogic$ formula. 

The function $\mathrm{trans} : \VRSLogic \to \FOformulas$, mapping $\VRSLogic$ formulae to formulae in first-order standpoint logic, is recursively defined as follows:\\[-1ex]
\noindent
\parbox{\textwidth}{
\begin{align*}
    &\trans{\pf{K}(\tau)} = \pf{K}(\tau)&\trans{\n\phi} = \n \trans{\phi} &\\
    &\trans{\pf{A}(\vec\tau)} = \pf{A}(\vec\tau)                       &\trans{\phi_1\! \con\! \phi_2}  = \trans{\phi_1} \wedge \trans{\phi_2}& \\  
    &\trans{\A x\, \phi(x)} = \forall x\ (\pf{ink}(x) \rightarrow \trans{\phi}) &\trans{\standbs\phi}  = \standbs \trans{\phi}&\\
\end{align*}
\vspace{-10.7ex}
\begin{gather*}
\trans{\pf{Q}(\tau_{1},\mydots,\tau_{k})} = 
\exists e_{1},\mydots,e_{k}(\pf{Q}(e_{1},\mydots,e_{k}) 
\land \pf{prec}(\tau_{1},e_{1})) \con \mydots \con \pf{prec}(\tau_{k},e_{k}))\ 
\end{gather*}
}

\noindent Then, we consider the set $\Phi$ of axioms in $\FOformulas$ such that:
\begin{align}
   \Phi=&\{\ \forall x (\pf{ind}(x) \leftrightarrow \neg\pf{ext}(x)),& 
    &\forall x (\pf{ind}(x)\! \leftrightarrow \!\Box_{*}\pf{ind}(x))\! \land\! (\pf{ext}(x)\! \leftrightarrow\! \Box_{*}\pf{ext}(x)), \\  
    &\bigwedge\textstyle_{\scaleobj{1}{\mathsf{Q}\in\mathcal{Q}}}\forall \vec x ( \pf{Q}(\vec x) \leftrightarrow \Box_{*}\pf{Q}(\vec x) ), &
    &\forall x,y (\pf{prec}(x,y) \rightarrow (\pf{ind}(x) \land \pf{ext}(y))),\ \\
    &\forall x \exists y (\pf{ind}(x)\!\rightarrow \!\pf{prec}(x,y) ), &
    &\forall x,y (\pf{prec}(x,y) \con \pf{prec}(x,z) \rightarrow y=z), \\
    &&&  \!\!\!\!\!\!\! 
    \forall x_1,..,x_k (\pf{A}(x_1,..,x_k)\! \rightarrow\! \pf{ink}(x_1)\! \con\!\mydots\! \con\!\pf{ink}(x_k) ),\\
    &&&  \!\!\!\!\!\!\! 
    \forall x \ (\pf{ink}(x)\! \rightarrow\! \pf{ind}(x)) \wedge ({\pf{ink}}(x)\! \leftrightarrow \! \bigvee\textstyle_{\scaleobj{0.75}{\mathsf{K}\in\mathcal{K}}} \pf{K}(x)) \ \}
\end{align}

These axioms state that every domain element is either an individual or an extension but not both, with $\pf{ind}$ and $\pf{ext}$ rigid predicates (line 1); that every precise entity property predicate is rigid; that $\pf{prec}(x,y)$ relates an individual $\pf{ind}(x)$ to a precise entity $\pf{ext}(y)$ (line 2), and that every individual $\pf{ind}(x)$ has (at each precisification) a unique precise extension $y$, $\pf{prec}(x,y)$ (line 3); that every element of a domain with an indefinite predicate $\pf{A}$ must be an individual in that precisification ($\pf{ink}$) (line 4), and finally that an individual in that precisification is an individual of some sort $\mathsf{k}\in\mathsf{K}$ (line 5). 

Finally, for a $\VRSLogic$ formula $\psi$ , we specify:
$$
\mathrm{Trans}(\psi):=\trans{\psi}\wedge\bigwedge_{\phi\in\Phi}\allstandb\phi$$
A $\VRSLogic$ formula $\psi$ and its translation  $\mathrm{Trans}(\psi)$ are equisatisatisfiable, that is, $\psi$ is $\VRSLogic$-satisfiable if and only if $\mathrm{Trans}(\psi)$ is $\FOformulas$-satisfiable. Correctness of the translation can be shown by establishing a correspondence between the two kinds of model in each direction. The proof is by induction and is straightforward, yet lengthy, thus it is omitted in the current manuscript.

It is worth remarking that the translation is linear in the size of the formula and preserves the monodic fragment of standpoint logic, where modalities occur only in front of formulas with at most one free variable. The monodic fragment of first-order modal logic is known to be decidable \cite{gabbay03many-dimensional}, and tight complexity bounds have been shown for monodic fragments of standpoint logic \cite{Gomez23StandpointEL}. This highlights the usefulness of the translation, since reasoning support for $\VRSLogic$ can be provided through its implementation.

\def\pr#1{\hbox{$\mathsf{#1}$}}


\section{Some Philosophical Views Considered in relation to the $\VRSLogic$ Semantics}
\label{SEC:phil-views}

From the translation, we may observe that we move from having a domain with only precise entities to also having a potentially large collection of \emph{indefinite entities} that are abstract in nature and only relate to their extension via the predicate $\pf{prec}$. 
This issue regarding whether multiple precise versions of vague things can  exist simultaneously (especially fuzzy boundaried objects like clouds) is known to philosophers as `The Problem of the Many'
\cite{Unger80many}, and has been the subject of considerable debate \cite{Weatherson09sep-short}.

One view is that the issue can be resolved by modifying the identity relation. 
According to Geach \cite{Geach80r+g}, the different precise versions of a cat are identical relative to the
high-level count noun `cat', whereas Lewis \cite{Lewis93a} suggests that physical objects that are almost the same
with respect to spatial extension and material composition are ordinarily regarded as identical. Our semantics has some affinity
with Geach's approach in that we consider that it is count nouns that determine individuation and hence the sets of
possible precise versions of individuals. However, in our theory there is no need to modify identity, since quantification
is over individuals not their possible precise extensions. 
We believe that Lewis's identity condition cannot account for many examples of objects with vague boundaries, where there
can be large difference between what we may take to be their spatial extension. For instance, small variations in our interpretation of `desert' may in certain circumstances result in very different spatial extensions being ascribed.

Lewis, along with Varzi \cite{Varzi01a}, also suggests a supervaluationist account where the 
different candidate extensions of a vague object correspond to their extension in different precisifications.
So, at a given precisification
there is always a single physical extension of each object.
But what is distinctive about our semantics is that we model variability in the extension of an individual
as an additional kind of vagueness, not subsumed by the indeterminacy of a name (or sortal) in referring to
an individual (or class of individuals).

Lowe \cite{Lowe82a} posits an ontological difference between objects of a high-level category such as `cat' and physical portions of feline tissue present in the world, which he calls `cat-constituters'. This is very much in line with our semantics, especially as presented in the flat (translated) form that contains both definite and indefinite
basic elements. Critics of Lowe (e.g.\ Lewis) have argued that the existence of both cats and cat-constituters is ontologically profligate (or incoherent).
We consider instances of the high-level categories to be semantic objects modelling the way language operates rather than carrying ontological commitment.
This is clearest in the semantics of $\VRSLogic$, where modelling indefinite objects as functions from precisifications to extensions 
is a way of packaging indeterminacy within a semantic object. In the flat version of our semantics, indefinite objects are present (along with precise extensions)
as basic elements of the domain; however we still consider them semantic rather than {\em ontic\/} elements.




\section{Conclusions and Further Work}



We have analysed the phenomenon of vagueness in both predicates and  objects, and  considered examples suggesting that the aspects of individuation, classification and demarcation associated with predication  need to be
considered separately. 
We have proposed the framework of \emph{variable reference logic}, which takes the general form of a supervaluation semantics but allows variability both in the set of individuals falling under a predicate and also 
in the spatial extension and physical constitution of `objects' denoted by names
or quantified variables. This separation not only supports flexible expressive capabilities of our formal language, but enables the representation to capture definite aspects of information despite other aspects being affected by vagueness. 

There remains much interesting work to be done investigating what can be expressed
and inferred by means of $\VRSLogic$.
We envisage reasoning support for $\VRSLogic$ being of practical interest for specific kinds of application, such as querying of geographical information systems,
and plan to develop a case study in this area.
The linear translation to flat standpoint logic and ongoing progress on implementations for that logic are a strong indicator that query answering functionality may be feasible.

\vspace{-0.7ex}


\bibliography{huge,newbrandon,local}

\appendix

\section{Proofs}

\begin{proof}[Translation Correctness Proof]

We now show that for a formula $\psi$ in $\VRSLogic$, $\VRSLogic$-satisfiability of $\psi$ coincides with $\FOformulas$-satisfiability of $\Trans{\psi}$.

($\Rightarrow$) Assuming the satisfiability of $\psi$ with $\VRSLogic$-vocabulary $\mathcal{V}^{\smallv}= \tuple{ \mathcal{K}, \mathcal{A}, \mathcal{Q},
  \mathcal{N}, \mathcal{S} }$, consider some $\VRSLogic$ model $\model^{\smallv}=\tuple{ E^{\smallv}, \Pi^{\smallv}, \sigma^{\smallv}, \delta^{\smallv}}$ of it. 
  
  Let now $\model=\tuple{ \Delta, \Pi, \sigma, \delta}$ denote the $\FOformulas$-interpretation for the vocabulary $\mathcal{V}= \tuple{ \mathcal{K}\cup \mathcal{A}\cup \mathcal{Q}, \mathcal{N}, \mathcal{S} }$ such that:
  \begin{enumerate}[label={(I.\arabic*)}]
      \item $\Delta = E^{\smallv}\cup I_{*}$, i.e. the domain is the union of the precise entities and the indefinite individuals,
      \item $\Pi=\Pi^v$ and $\sigma=\sigma^v$, i.e. the set of precisifications and standpoint structure are as in $\model^{\smallv}$,
      \item the interpretation function $\delta$ is such that 
      \begin{enumerate}[label={(I.3.\arabic*)}]
          \item $\delta^{\smallv}_{\mathcal{K}} \cup \delta^{\smallv}_{\mathcal{A}} \cup \delta^{\smallv}_{\mathcal{N}}\subseteq\delta$,
          \item $\interpret{\pf{Q}}{\delta(\pi)}=\interpret{\pf{Q}}{\delta^{\smallv}_{\mathcal{Q}}}$ for each $\pi\in\Pi$ and $\pf{Q}\in\mathcal{Q}$,
          \item for each $\pi\in\Pi$, 
          \begin{enumerate}[label={(I.3.3.\arabic*)}]
              \item $\interpret{\pf{ext}}{\delta(\pi)}=E^{\smallv}$,
              \item  $\interpret{\pf{ind}}{\delta(\pi)}=I_{*}$,
              \item  $\interpret{\pf{ink}}{\delta(\pi)}=I_{\pi}$, and
              \item  $\interpret{\pf{prec}}{\delta(\pi)}=\{(i,e)\mid i\in I_{*}, i(\pi)=e\}$.
          \end{enumerate}
      \end{enumerate}
  \end{enumerate}
  
 First, we show that $\model\models \bigwedge_{\phi\in\Phi}\allstandb\phi$ by construction, with $\Phi$ the set of axioms defined in the translation.
  \begin{enumerate}[label={(\arabic*)}]
     \item [(1.a):] Notice that $\Delta= E^{\smallv}\cup I_{*}$ and clearly $ E^{\smallv}\cap I_{*}=\emptyset$, hence by construction (I.3.3.1 and I.3.3.2) we have that for $d\in\Delta$ then $d\in\pf{ind}^{\delta(\pi)}\cup\pf{ext}^{\delta(\pi)}$ and $\pf{ind}^{\delta(\pi)}\cap\pf{ext}^{\delta(\pi)}=\emptyset$ for all $\pi\in\Pi$, thus $\model,\pi\models \forall x (\pf{ind}(x) \leftrightarrow \neg\pf{ext}(x))$ and finally $\model\models\allstandb \forall x (\pf{ind}(x) \leftrightarrow \neg\pf{ext}(x))$ as desired. 
     \item [(1.b):] From (I.3.3.2)  if $d\in\interpret{\pf{ind}}{\delta(\pi)}$ then for all $\pi'\in\Pi$ also $d\in\interpret{\pf{ind}}{\delta(\pi')}$ and from (I.3.3.1) $d\in\interpret{\pf{ext}}{\delta(\pi)}$ then for all $\pi'\in\Pi$ also $d\in\interpret{\pf{ext}}{\delta(\pi')}$, hence by the semantics $\model,\pi\models \forall x (\pf{ind}(x)\! \leftrightarrow \!\Box_{*}\pf{ind}(x))\! \land\! (\pf{ext}(x)\! \leftrightarrow\! \Box_{*}\pf{ext}(x))$ and thus $\model\models \allstandb\forall x (\pf{ind}(x)\! \leftrightarrow \!\Box_{*}\pf{ind}(x))\! \land\! (\pf{ext}(x)\! \leftrightarrow\! \Box_{*}\pf{ext}(x))$ as desired;
     \item [(2.a):] It follows from (I.3.2) that if $\vec d\in\interpret{\pf{Q}}{\delta(\pi)}$ then for all $\pi'\in\Pi$ also $\vec d\in\interpret{\pf{Q}}{\delta(\pi')}$, hence by the semantics $\model,\pi\models \bigwedge\textstyle_{\scaleobj{1}{\mathsf{Q}\in\mathcal{Q}}}\forall \vec x ( \pf{Q}(\vec x) \leftrightarrow \Box_{*}\pf{Q}(\vec x) )$, and thus $\model\models \allstandb\bigwedge\textstyle_{\scaleobj{1}{\mathsf{Q}\in\mathcal{Q}}}\forall \vec x ( \pf{Q}(\vec x) \leftrightarrow \Box_{*}\pf{Q}(\vec x) )$ as desired;
     \item [(2.b):] by construction from (I.3.3.4) we have that $\interpret{\pf{prec}}{\delta(\pi)}=\{(i,e)\mid i\in I_{*}, i(\pi)=e\}$, and by the semantics $i(\pi)\in E^{\smallv}$. Together with (I.3.3.1) and (I.3.3.2), we clearly have $\model,\pi\models\forall x,y (\pf{prec}(x,y) \rightarrow (\pf{ind}(x) \land \pf{ext}(y)))$, and thus  $\model\models\allstandb\forall x,y (\pf{prec}(x,y) \rightarrow (\pf{ind}(x) \land \pf{ext}(y)))$;
     \item [(3.a), (3.b):] by construction from (I.3.3.2) we have that if $i\in\interpret{\pf{ind}}{\delta(\pi)}$ for some $\pi\in\Pi$ then $i\in I_{*}$, and from (I.3.3.4), we then have that  $(i,i(\pi))\in\interpret{\pf{prec}}{\delta(\pi)}$, hence clearly $\model\models\forall x \exists y (\pf{ind}(x)\!\rightarrow \!\pf{prec}(x,y) )$. And since for each $\pi\in\Pi$ and $i\in I_{*}$ only $(i,i(\pi))\in\interpret{\pf{prec}}{\delta(\pi)}$, then also $\model,\pi\models\forall x,y (\pf{prec}(x,y) \con \pf{prec}(x,z) \rightarrow y=z)$ and this $\model\models\allstandb(\forall x,y (\pf{prec}(x,y) \con \pf{prec}(x,z) \rightarrow y=z))$ as desired;
     \item [(4):] for $\pf{A}$ of arity $k$, if $\vec d \in \interpret{\pf{A}}{\delta(\pi)}$ then by construction (I.3.1) and by the semantics we have $\vec d \in I_{\pi}^k$. From (I.3.3.3) we have that for all $d \in I_{\pi}$ also $d\in\interpret{\pf{ink}}{\delta(\pi)}$  and hence $\model\models\allstandb(\forall x_1,..,x_k (\pf{A}(x_1,..,x_k)\! \rightarrow\! \pf{ink}(x_1)\! \con\!\mydots\! \con\!\pf{ink}(x_k) ))$ as desired;
     \item [(5):] Finally, by construction we have that if  $d\in\interpret{\pf{ink}}{\delta(\pi)}$ then $d \in I_{\pi}$, and thus by the semantics both $d \in I_{*}$ and $d\in\bigcup \{ \interpret{\mathsf{K}}{{\delta_{\mathcal{K}}(\pi)}} \ |
  \ \mathsf{K} \in \mathcal{K}\}$. Then from (I.3.1), (I.3.3.2) and (I.3.3.3) we obtain $\model\models\allstandb(\forall x \ (\pf{ink}(x)\! \rightarrow\! \pf{ind}(x)) \wedge ({\pf{ink}}(x)\! \leftrightarrow \! \bigvee\textstyle_{\scaleobj{0.75}{\mathsf{K}\in\mathcal{K}}} \pf{K}(x)))$ as desired;
 \end{enumerate}
  
 We now show by structural induction on $\psi$ that, for every $\pi\in\Precs$, if $\model^{\smallv}, \pi, v^{\smallv} \models \psi$ then $\model, \pi, v \models \mathrm{trans}(\psi)$ for $v$ an extension of $v^{\smallv}$. 
 
 \begin{description}
 \item [$\psi = \pf{K}(\tau)$:] Assume $\model^{\smallv}, \pi, v^{\smallv} \models \pf{K}(\tau)$. Then we show that $\model, \pi, v \models \mathrm{trans}(\pf{K}(\tau))$ for $v$ an extension of $v^{\smallv}$. Notice that $\mathrm{trans}(\pf{K}(\tau))=\pf{ind}(\tau) \land \pf{K}(\tau)$. 
 By the semantics of $\VRSLogic$ we have $\mathsf{K}^{\delta^{\smallv}_{\mathcal{K}}(\pi)} \subseteq I_{*}$ and $\tau^{\pi, v^{\smallv}}\in \mathsf{K}^{\delta^{\smallv}_{\mathcal{K}}(\pi)}$. Then, since $\pf{ind}^{\delta(\pi)}=I_{*}$ for all $\pi\in\Pi$, and $v$ is an extension of $v^{\smallv}$, then $\tau^{\pi, v}\in \pf{ind}^{\delta(\pi)}$ for all $\pi\in\Pi$.
 Similarly, since $\delta^{\smallv}_{\mathcal{K}}\subseteq\delta$ and $v$ is an extension of $v^{\smallv}$, then $t^{\pi, v}\in \pf{K}^{\delta(\pi)}$ and hence $\model,\pi, v\models\pf{ind}(\tau) \land \pf{K}(\tau) $.
 
 \item [$\psi = \pf{A}(\tau)$:] Proceeds similarly to the previous case.
 \item [$\psi = \pf{Q}(\tau)$:] Assume $\model^{\smallv}, \pi, v^{\smallv} \models \pf{Q}(\tau)$. Then we show that $\model, \pi, v \models \mathrm{trans}(\pf{Q}(\tau))$ for any $v$ that is an extension of $v^{\smallv}$. Notice that $\mathrm{trans}(\pf{Q}(\tau))=\pf{ind}(\tau) \land \exists x( \pf{Q}(x) \land \pf{prec}(\tau,x))$. 
     Let $\tau^{\pi, v^{\smallv}}=i$. By the semantics of $\VRSLogic$ we have $i\in I_{*}$ and $i(\pi)\in\mathsf{Q}^{\delta^{\smallv}_{\mathcal{Q}}}$, with $i(\pi)\in E^{\smallv}$. Then, since $\pf{ind}^{\delta(\pi)}=I_{*}$ for all $\pi\in\Pi$, and $v$ is an extension of $v^{\smallv}$, then $t^{\pi, v}\in \pf{ind}^{\delta(\pi)}$ for all $\pi\in\Pi$. 
     Now, from the translation we have $\model, \pi, v \models \exists x( \pf{Q}(x) \land \pf{prec}(\tau,x))$. By construction, for $i(\pi)=e$ we have that $e\in\pf{Q}^{\delta(\pi)}$ from (I.3.2) and $(i,e)\in\pf{prec}^{\delta(\pi)}$ from (I.3.3.4). And since $e\in E^{\smallv}$, then it is not mapped by $v^{\smallv}$, so there is a variable assignment $v_{[x\rightarrow e]}$ extending $v^{\smallv}$ and with $v(x)=e$. Consequently, $\model,\pi, v\models\pf{ind}(\tau) \land \exists x( \pf{Q}(x) \land \pf{prec}(\tau,x))$ as desired.
 \item [$\psi = \neg\psi'$:] Trivial.
 \item [$\psi = \psi'\land\psi''$:] Trivial.
 \item [$\psi = \A x\,\psi'(x)$:] Assume $\model^{\smallv}, \pi, v^{\smallv} \models \A x\,\psi'(x)$. Then we show that $\model, \pi, v \models \mathrm{trans}(\A x\,\psi'(x))$ for any $v$ that is an extension of $v^{\smallv}$. Notice that $\mathrm{trans}(\A x\,\psi'(x))=\A x\,(\pf{ink}(x) \imp \trans{\psi'})$. Hence we must show that for all $d\in\Delta$, if $d\in\pf{ink}^{\delta(\pi)}$ then $\model, \pi, v_{x\rightarrow d} \models \trans{\psi'}(x)$. By construction (I.3.3.3) we have $\pf{ink}^{\delta(\pi)}=I_{\pi}$, and by the assumption and the semantics of $\VRSLogic$ we have that $\model^{\smallv}, \pi, v^{\smallv}_{\set{x\mapsto i}} \models \A x\,\psi'(x)$ for all $i \in I_{\pi}$. With this and by the inductive hypothesis it is clear that $\model,\pi, v\models \mathrm{trans}(\A x\,\psi'(x))$
 
 
 \item [$\psi = \standb{\ste}\psi'$:] Trivial.
 \end{description}
 
 It is then a direct consequence that $\model$ is a model of both conjuncts of $\mathrm{Trans}(\phi)$. Hence $\model$ is a model of $\mathrm{Trans}(\phi)$, witnessing its satisfiability.
 \medskip
 
 ($\Leftarrow$)  Consider now a $\FOformulas$-model $\model^{\smalls}$ of $\mathrm{Trans}(\psi)$ with vocabulary $\mathcal{V}^{\smalls}= \tuple{ \mathcal{P},\mathcal{N}, \mathcal{S} }$. 
 Define the vocabulary $\mathcal{V}= \tuple{ \mathcal{K}, \mathcal{A}, \mathcal{Q},
  \mathcal{N}, \mathcal{S} }$, by letting
  \begin{itemize}
      \item $\mathcal{K}=\{\pf{P}\in\mathcal{P}\backslash \{\pf{ind}\} \mid \pf{P}^{\delta^{\smalls}(\pi)} \cap \pf{ind}^{\delta^{\smalls}(\pi)}\neq\emptyset\}$
      \item $\mathcal{A}=\emptyset$
      \item $\mathcal{Q}=\{\pf{P}\in\mathcal{P}\backslash \{\pf{ext}\} \mid \pf{P}^{\delta^{\smalls}(\pi)} \cap \pf{ext}^{\delta^{\smalls}(\pi)}\neq\emptyset\}$
  \end{itemize}
  
  We now define the $\VRSLogic$ model $\model=\tuple{ E, \Pi, \sigma, \delta}$ with vocabulary $\mathcal{V}$, by letting 
  
  \begin{enumerate}[label={(M\arabic*)}]
      \item $\Pi$ and $\sigma$ be as in $\model^{\smalls}$
      \item $E=\pf{ext}^{\delta^{\smalls}(\pi)}$ for some $\pi\in\Pi$ 
      \item $\idom=Maps(\Pi,E)$ is the set of all mappings from  precisifications to precise entities.  
      \item $I_{\pi}=\pf{ink}^{\delta^{\smalls}(\pi)}$, $\pf{K}^{\delta_\mathcal{K}(\pi)} = \pf{K}^{\delta^{\smalls}(\pi)}$ and $\pf{N}^{\delta_\mathcal{N}(\pi)} = \pf{N}^{\delta^{\smalls}(\pi)}$ for all $\pi \in \Pi$, $\pf{K}\in\mathcal{K}$ and $\pf{N}\in\mathcal{N}$,
      \item $\pf{Q}^{\delta_\mathcal{Q}} = \pf{Q}^{\delta^{\smalls}(\pi)}$ for any $\pi \in \Pi$ and all $\pf{Q}\in\mathcal{Q}$.
  \end{enumerate}
  
First, we show that the structure thus defined is indeed a $\VRSLogic$, recalling that $\model^{\smalls}\models \bigwedge_{\phi\in\Phi}\allstandb\phi$ with $\Phi$ the set of axioms defined in the translation.
By axiom (1.b) we have that for all $\pi,\pi'$, $\pf{ind}^{\delta^{\smalls}(\pi)}=\pf{ind}^{\delta^{\smalls}(\pi')}$ and $\pf{ext}^{\delta^{\smalls}(\pi)}=\pf{ext}^{\delta^{\smalls}(\pi')}$. For the sake of abbreviation we will refer to these sets as $\pf{ind}^{\smalls}$ and $\pf{ext}^{\smalls}$. Following, by (1.a) we have that $E^{\smalls}=\pf{ind}^{\smalls}\cup\pf{ext}^{\smalls}$ and $\pf{ind}^{\smalls}\cap\pf{ext}^{\smalls}=\emptyset$. Hence, the set of indefinite individuals and the set of precise entities are disjoint in $\model^{\smalls}$ and independent from $\Pi$. 
Moreover, by axiom (3.a) every individual $i\in\pf{ind}^{\smalls}$ has a precise extension at a given precisification $\pi$, i.e. there is some $e$ such that $(i,e)\in\pf{precs}^{\delta^{\smalls}(\pi)}$ and by (2.b) $e\in\pf{ext}^{\smalls}$. And, by (3.b) there is no other $e'$ such that $(i,e')\in\pf{precs}^{\delta^{\smalls}(\pi)}$ and hence we can set $i(\pi)=e$ as desired, thus obtaining from each $i\in\pf{ind}^{\smalls}$ an indefinite individual $i\in \idom$.
Finally, $I_{\pi}$, the set of indefinite individuals in $\pi$, corresponds to $\pf{ink}^{\delta^{\smalls}(\pi)}$ because by axiom (4.b) we have $\pf{ink}^{\delta^{\smalls}(\pi)}\subseteq\pf{ind}^{\smalls}$ and $\pf{ink}^{\delta^{\smalls}(\pi)}=\bigcup\textstyle_{\scaleobj{0.75}{\mathsf{K}\in\mathcal{K}}} \pf{K}^{\delta^{\smalls}(\pi)}$, which given that $\pf{K}^{\delta_\mathcal{K}(\pi)} = \pf{K}^{\delta^{\smalls}(\pi)}$, clearly encodes the definition of $I_{\pi}$.

From this we have that $\model$ is a $\VRSLogic$ structure and
 \begin{enumerate}[label={(P\arabic*)}]
      \item[A0] every $i\in \pf{ind}^{\smalls}$ corresponds to an indefinite individual $i\in\idom$ such that for each $\pi\in\Precs$ we have $i(\pi)=e$ if $(i,e)\in\pf{precs}^{\delta^{\smalls}(\pi)}$, and
     \item[A1] the set $I_{\pi}$ of indefinite individuals in $\pi$ corresponds to $\pf{ink}^{\delta^{\smalls}(\pi)}$.
  \end{enumerate}

 To show that $\model$ is a model of $\psi$, we show by structural induction over $\psi$ that, for every $\pi\in\Precs$ and variable assignment $v$, if $\model^{\smalls}, \pi, v^{\smalls} \models \trans{\psi}$ then $\model, \pi, v \models \psi$, with $v$ the restriction of $v^{\smalls}$ to the range of $\idom$. Thus, the established modelhood of $\model$ ensures satisfiability of $\phi$.  
  \begin{description}
     \item [$\trans{\pf{K}(\tau)}$] Assume that $\model^{\smallv}, \pi, v^{\smalls} \models \mathrm{trans}({\pf{K}(\tau)})$. Then $\model^{\smalls}, \pi, v^{\smalls} \models \pf{K}(\tau)$. Consequently, there is an element $i\in E^{\smalls}$ such that $i=\tau^{\pi,v^{\smalls}}$, $i\in\pf{K}^{\delta^{\smalls}(\pi)}$, and by axiom (4.b) $i\in\pf{ind}^{\smalls}$. By [A0], $i$ translates to an indefinite individual in $\idom$. By the axiom (4.b) we have that for all $\pi$, $\pf{K}^{\delta^{\smalls}(\pi)}\subseteq\pf{ind}^{\smalls}$ and thus again by [A0], $\pf{K}^{\delta^{\smalls}(\pi)}$ translates to a set of indefinite individuals $\pf{K}^{\delta_\mathcal{K}(\pi)}\subseteq\idom$ such that $i\in\pf{K}^{\delta_\mathcal{K}(\pi)}$ as desired.
     \item [$\trans{\pf{A}(\tau)}$] Assume that $\model^{\smalls}, \pi, v^{\smalls} \models \mathrm{trans}({\pf{A}(\tau)})$. Then $\model^{\smalls}, \pi, v^{\smalls} \models \pf{A}(\tau)$. Consequently, there is an element $i\in E^{\smalls}$ such that $i=\tau^{\pi,v^{\smalls}}$, $i\in\pf{A}^{\delta^{\smalls}(\pi)}$, and by axioms (4.a, 4.b) $i\in\pf{ind}^{\smalls}$. By [A0], $i$ translates to an indefinite individual in $\idom$. By axioms (4.a,4.b) it is clear that for all $\pi$ we have $\pf{A}^{\delta^{\smalls}(\pi)}\subseteq\pf{ink}^{\delta^{\smalls}(\pi)}$ and $\pf{ink}^{\delta^{\smalls}(\pi)}\subseteq\pf{ind}^{\smalls}$. Thus, again by [A0], $\pf{A}^{\delta^{\smalls}(\pi)}$ translates to a set of indefinite individuals $\pf{A}^{\delta_\mathcal{A}(\pi)}\subseteq I_{\pi}$ such that $i\in\pf{A}^{\delta_\mathcal{A}(\pi)}$ as desired.
     \item [$\trans{\pf{Q}(\tau)}$] Assume that $\model^{\smalls}, \pi, v^{\smalls} \models \mathrm{trans}({\pf{Q}(\tau)})$. 
     Then $\model^{\smalls}, \pi, v^{\smalls} \models \exists x( \pf{Q}(x) \land \pf{prec}(\tau,x))$.
     Consequently, there are $i,e\in E^{\smalls}$ such that $i=\tau^{\pi,v^{\smalls}}$, 
     $e=v^{\smalls}(x)$, $(i,e)\in\pf{prec}^{\delta^{\smalls}(\pi)}$, 
     $e\in\pf{Q}^{\delta^{\smalls}(\pi)}$, 
     and by axiom (2.b) $i\in\pf{ind}^{\smalls}$ and $e\in\pf{ext}^{\smalls}$. 
     By [A0], $i$ translates to an indefinite individual in $\idom$ such that $i(\pi)=e$. 
     Also, from axiom (2.a) we have that for all $\pi,\pi'$, $\pf{Q}^{\delta^{\smalls}(\pi)}=\pf{Q}^{\delta^{\smalls}(\pi')}$, and we recall that  $\pf{Q}^{\delta^{\smalls}(\pi')}=\pf{Q}^{\delta_{\mathcal{Q}}}$ for some $\pi'\in\Pi$. 
     Finally, since $v$ is a restriction on $v^{\smalls}$ to the range of $\idom$, 
     then $\interpret{\tau}{\pi,\varassign}=\interpret{\tau}{\pi,\varassign^{\smalls}}$ 
     and consequently $\interpret{\tau}{\pi,\varassign}(\pi) \in \pf{Q}^{\deltaQ}$. 
     Thus $\kstruct,\!\pi,\!\varassign  \models \pf{Q}(\tau)$ as desired.
     \item [$\trans{{\neg \psi}}$] Trivial.
     \item [$\trans{\psi\wedge\psi'}$] Trivial.
     \item [$\trans{\standbs\psi}$] Trivial.
     \item [$\trans{\A x \psi}$] Assume that $\model^{\smalls}, \pi, v^{\smalls} \models \mathrm{trans}({\A x \psi})$. Then $\model^{\smalls}, \pi, v^{\smalls} \models \A x (\pf{ink}(x) \rightarrow \trans{\psi})$. Then, by the semantics we have $\kstruct,\!\pi,\!\varassign\models  (\pf{ink}(x) \rightarrow \trans{\psi}) \text{ for all } \varassign:\Vars\to\Dom$. This means that for all $i\in\Delta$ if $i\in\pf{ink}^{\delta^{\smalls}(\pi)}$ then $\model^{\smalls}, \pi, v^{\smalls} \models \trans{\psi}$ with $v(x)=i$. By the inductive hypothesis we have that  $\kstruct,\!\pi,\!\varassign \models \psi(x)$ for $v$ the restriction on $v^{\smalls}$ over the range of $\idom$. Then since by [A1] we have that $I_{\pi}=\pf{ink}^{\delta^{\smalls}(\pi)}$, we obtain that $\kstruct,\!\pi,\!\varassign_{\set{x\mapsto i}} \models\  \psi(x)\hbox{~~for all}\ i \in I_\pi$, and thus $\kstruct,\!\pi,\!\varassign_{\set{x\mapsto i}} \models\  \A x \psi$ as desired.
 \end{description}

\end{proof}

\end{document}